\documentclass{bmvc2k}

\usepackage{multirow}
\usepackage{pifont}%
\usepackage{amsmath} 
\usepackage{amsfonts} 
\usepackage{amssymb} 
\usepackage{tabularx}
\usepackage{float}

\newcommand{\cmark}{\ding{51}}%
\newcommand{\xmark}{\ding{55}}%

\title{RT-GS2: Real-Time Generalizable Semantic Segmentation for 3D Gaussian Representations of Radiance Fields}

\addauthor{Mihnea-Bogdan Jurca*}{mihnea-bogdan.jurca@vub.be}{1, 2}
\addauthor{Remco Royen*}{remco.royen@vub.be}{1}
\addauthor{Ion Giosan}{ion.giosan@cs.utcluj.ro}{2}
\addauthor{Adrian Munteanu}{adrian.munteanu@vub.be}{1}

\addinstitution{
 Department ETRO\\
 Vrije Universiteit Brussel\\
 Brussels, Belgium
}
\addinstitution{
 Computer Science Department\\
 Technical University of Cluj-Napoca,\\
 Cluj-Napoca, Romania
}

\runninghead{Jurca, Royen, Giosan, Munteanu}{RT-GS2}

\begin{document}

\maketitle

\begin{abstract}


Gaussian Splatting has revolutionized the world of novel view synthesis by achieving high rendering performance in real-time. Recently, studies have focused on enriching these 3D representations with semantic information for downstream tasks. In this paper, we introduce RT-GS2, the first generalizable semantic segmentation method employing Gaussian Splatting. While existing Gaussian Splatting-based approaches rely on scene-specific training, RT-GS2 demonstrates the ability to generalize to unseen scenes. Our method adopts a new approach by first extracting view-independent 3D Gaussian features in a self-supervised manner, followed by a novel View-Dependent / View-Independent (VDVI) feature fusion to enhance semantic consistency over different views. Extensive experimentation on three different datasets showcases RT-GS2's superiority over the state-of-the-art methods in semantic segmentation quality, exemplified by a 8.01\% increase in mIoU on the Replica dataset. Moreover, our method achieves real-time performance of 27.03 FPS, marking an astonishing 901 times speedup compared to existing approaches. This work represents a significant advancement in the field by introducing, to the best of our knowledge, the first real-time generalizable semantic segmentation method for 3D Gaussian representations of radiance fields. The project page and implementation can be found at \url{https://mbjurca.github.io/rt-gs2/} 

\end{abstract}


\section{Introduction}
\label{sec:intro}

Scene Understanding is a fundamental area of research, essential for facilitating seamless interactions between digital devices and the three-dimensional environment. While 2D representations such as RGB images are traditionally being employed, they fail to fully capture the three-dimensional properties of the scene, resulting in view-dependent outcomes~\cite{di2024deepkalpose}. On the other hand, 3D representations such as point clouds and polygonal meshes offer the capability to digitally represent 3D scenes, but typically require high-resolution 3D point data to capture fine details~\cite{vu2022softgroup, royen2023resscal3d}, often obtained through expensive devices like LiDARs.


In recent years, Neural Radiance Fields (NeRFs) have emerged as a groundbreaking approach for novel view synthesis~\cite{mildenhall2021nerf, barron2022mip, muller2022instant}. By training multi-layered perceptrons (MLPs) on extensive sets of images captured from various viewpoints, NeRFs learn an implicit 3D representation of a scene. Beyond novel view synthesis, this learned 3D representation proves beneficial for downstream tasks as it enhances accuracy and enables view-consistent scene understanding \cite{wang2022clip}. Moreover, by formulating the loss of rendering and downstream tasks in the 2D domain, the need for time-intensive 3D annotations is eliminated. However, despite their performance, NeRFs exhibit a significant trade-off between visual quality and inference speed \cite{muller2022instant, yu2021plenoctrees}. Recently, a novel approach, dubbed Gaussian Splatting, was introduced for novel view synthesis \cite{kerbl20233d}. This method learns 3D Gaussian distributions in space, encompassing not only location and scale but also opacity and color spherical harmonics per 3D Gaussian. By simply splatting these Gaussians during inference, real-time high-quality synthesis of novel views is achieved.

\begin{figure}
    \centering
    \subfigure{\includegraphics[width=1\textwidth]{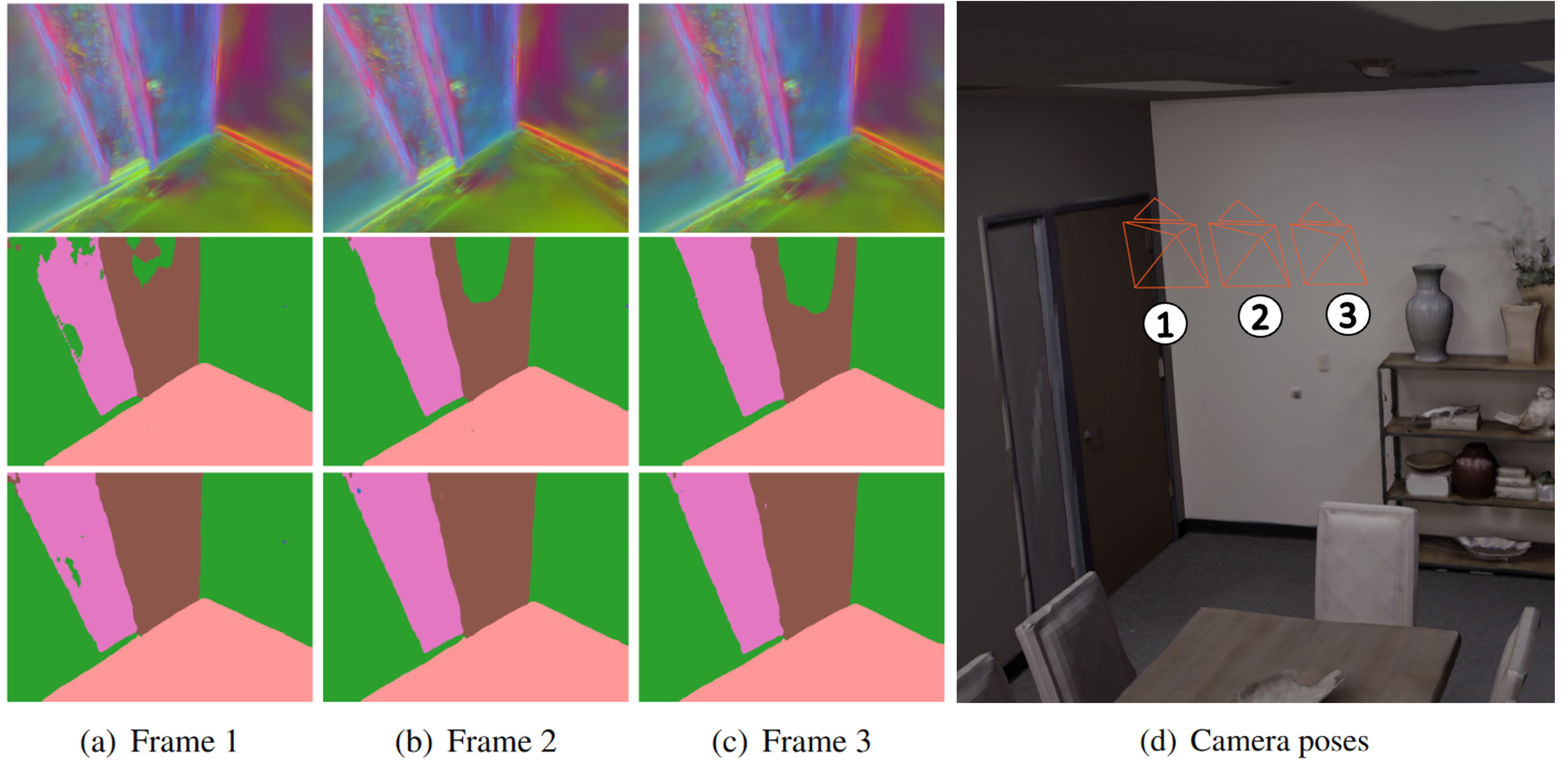}} 
    \caption{Visualization of the enhanced view-consistency throughout subsequent frames. (Top) Visualization of our view-independent 3D features using PCA, (middle) semantic segmentation without the usage of view-independent 3D features, and (bottom) proposed semantic segmentation when using our view-independent 3D features.}
    \label{fig:3D_feat}
\end{figure}


Significant research efforts have been directed towards extending novel view synthesis methods to accommodate downstream tasks. A popular approach is to equip NeRFs with semantic segmentation capabilities~\cite{zhi2021place, cen2023segment}. While these methods yield impressive segmentation results, they are trained on a scene-per-scene basis and are thus unable to generalize to unseen scenes. Consequently, methods~\cite{vora2021nesf, johari2022geonerf, liu2022neural, liu2023semantic, li2023gp, chou2024gsnerf} capable of performing semantic segmentation on unseen 3D NeRF representations were designed. As these methods are built upon NeRFs, real-time capabilities are lacking. Although recent studies have begun exploring downstream applications using 3D Gaussian Splatting~\cite{zhou2023feature, ye2023gaussian, qin2023langsplat, hu2024semantic}, to the best of our knowledge, no existing method in the current literature addresses generalizable semantic segmentation for 3D Gaussian Splatting.

In this paper, we present a novel method, dubbed RT-GS2, designed for generalizable semantic segmentation based on 3D Gaussian Splatting. RT-GS2 consists of three distinct stages. Firstly, a self-supervised 3D Gaussian feature extractor learns view-independent 3D features from the complete Gaussian 3D representation. After splatting the features and 3D Gaussians for a specific viewpoint, 
feature fusion is performed, enhancing view-consistent semantic segmentation, as illustrated in the examples of Figure \ref{fig:3D_feat}. Results for both semantic segmentation and depth prediction showcase the robustness of the obtained geometric 3D features. RT-GS2 not only surpasses the state of the art in generalizable semantic segmentation but also achieves an impressive speedup, being 901 times faster than existing methods. By doing so, it is the first to meet real-time constraints, marking a significant advancement in the field. In summary, our main contributions include:
\begin{itemize}
    \item The introduction of a novel method for generalizable semantic segmentation, the first to employ 3D Gaussian splatting. RT-GS2 presents a novel approach by first obtaining view-independent 3D features in a self-supervised manner, followed by a so-called View-Dependent / View-Independent (VDVI) feature fusion to obtain enhanced view-consistency for semantic segmentation.
    \item A self-supervised feature extractor for 3D Gaussians, enabling the extraction of generic and consistent view-independent 3D features, which prove to be robust for multiple tasks such as semantic segmentation and depth predictions.
    \item Extensive experimentation on different datasets demonstrating that RT-GS2 not only strongly outperforms the state of the art in segmentation quality but also achieves real-time inference, achieving a notable 901 times speedup compared to existing generalizable semantic segmentation methods.  
\end{itemize}

\section{Related work}

\noindent \textbf{Novel View Synthesis.}
The growing interest in implicit neural representations has greatly advanced the frontier of novel view synthesis in recent years. The seminal NeRF paper \cite{mildenhall2021nerf} has spurred iterative enhancements focusing on faster rendering \cite{reiser2021kilonerf, yu2021plenoctrees, wang2022r2l}, accelerated training processes \cite{sun2022direct, muller2022instant, chen2022tensorf}, and the capability to handle unbounded scenes \cite{barron2022mip}. Recently, Gaussian Splatting \cite{kerbl20233d} has demonstrated superiority over NeRF-based methods in both rendering quality and inference time. Acknowledging the remarkable potential of 3D Gaussians, we build upon this paradigm.




\noindent \textbf{Semantic Segmentation.} 
Traditional semantic segmentation techniques operate on a single modality. While 2D-based techniques \cite{ wang2023internimage, cheng2022masked} benefit from the employment of cost-efficient RGB-camera's, they have difficulties to fully capture the underlying 3D geometry. 3D semantic segmentation methods~\cite{qi2017pointnet, qi2017pointnet++, wu2019pointconv, zhao2021point, wu2023point} on the other hand, achieve high performance but require dense 3D models captured by expensive 3D scanners. The advent of NeRFs allowed to achieve view-consistent results on 2D images for a specific 3D scene by equipping NeRFs with semantic capabilities~\cite{zhi2021place, cen2023segment}. In order to achieve real-time constraints, \cite{ye2023gaussian, dou2024cosseggaussians, silva2024contrastive} learn semantic features for each 3D Gaussian.




\noindent \textbf{Generalizable Semantic Radiance Fields.}
While the above mentioned NeRF- and Gaussian Splatting-based semantic segmentation papers achieve high performance, their scene-specific training leads to significant overfitting for individual scenes. To address this limitation, generalizable semantic segmentation methods, capable of segmenting unseen scenes, were proposed for NeRFs ~\cite{liu2023semantic, li2023gp, chen2024gnesf, chou2024gsnerf}. For instance, S-Ray~\cite{liu2023semantic} introduces a Cross-Reprojection Attention module for efficient exploitation of semantic information along rays, while GP-NeRF \cite{li2023gp} utilizes transformers to aggregate semantic embedding fields. GNeSF~\cite{chen2024gnesf} employs a soft voting mechanism to aggregate 2D semantic information from different views, and GSNeRF \cite{chou2024gsnerf} integrates image semantics into the synthesis process for mutual enhancement. However, due to their dependence and build-up on NeRFs, these methods lack real-time execution. To our knowledge, there is no existing method in scientific literature that uses 3D Gaussian Splatting for real-time generalizable semantic segmentation. This is addressed next.




\section{3D Scene Representations Using 3D Gaussian Splats}

In order to render novel views of complex scenes, NeRF-based methods define a continuous volumetric radiance field~\cite{mildenhall2021nerf} for each scene, parameterized by $\mathbf{F}_\theta: \mathbb{R}^5 \rightarrow \mathbb{R}^4$, where $\mathbf{F}_\mathbf{\theta}$ is implemented by a MLP with learnable parameters $\mathbf{\theta}$. A differentiable forward mapping function is employed to retrieve discrete 2D views from the radiance field. A per-pixel loss between the synthesized rendering and the ground truth image allows the optimization of parameters $\theta$ for a specific scene.

Gaussian Splatting~\cite{kerbl20233d} takes a different approach and optimizes the training and rendering process while preserving the desirable properties of a radiance field. This is achieved by removing the implicit 3D representation and instead modeling each scene $k$ as a combination of 3D Gaussian functions $\mathbf{G}^{k} = \{\mathbf{g}^{k}_{1}, \mathbf{g}^{k}_{2}, \ldots, \mathbf{g}^{k}_{N}\}$. Each Gaussian $\mathbf{g}^{k}_{i}$ is defined by its world coordinates \( \mathbf{x}^k_i \in \mathbb{R}^3 \) and a covariance matrix \( \Sigma_i^k \in \mathbb{R}^{3 \times 3}\). Additionally, they are further enriched with an opacity \( \alpha_i^k \in \mathbb{R} \), and a color \( \mathbf{c}^k_i\), represented by spherical harmonic coefficients with three degrees. Thus, mathematically, each 3D Gaussian can be represented as $\mathbf{g}^{k}_i = \{\mathbf{x}^k_i, \Sigma_i^k, \alpha_i^k, \mathbf{c}^k_i\}$. To render from $\mathbf{G}^{k}$ the 2D image $\hat{\mathbf{I}}^k_j$ for view $j$ with pose $\mathbf{p}_j$, the 3D Gaussians are splatted into a RGB image employing alpha blending, expressed mathematically as follows:
\begin{equation}
\hat{\mathbf{I}}^k_j(u,v) = \sum_{i \in N'} \mathbf{c}_i^k \alpha_i^k \prod_{m=1}^{i-1} (1 - \alpha_m^k),
\label{eq:alphablendcolor}
\end{equation}
where $(u,v) \in ([1,H], [1,W])$ represent the pixel coordinates in the 2D image after splatting $N'$ Gaussians. When repeated for all pixels, the rendered image $\hat{\mathbf{I}}^k_j$ is obtained. The Gaussian parameters are optimized by employing the following loss function:
\begin{equation}
\mathcal{L} = (1 - \lambda) \mathcal{L}_1 + \lambda \mathcal{L}_{D-SSIM},
\label{eq:gsloss}
\end{equation}
with $\mathcal{L}_1$ a per-pixel $L_1$-loss and $\mathcal{L}_{D-SSIM}$ the structural dissimilarity metric between the rendered $\hat{\mathbf{I}}^k_j$ and the ground truth image $\mathbf{I}^k_j$. The hyperparameter $\lambda$ is typically set to 0.2. Similar to Equation \ref{eq:alphablendcolor}, a feature rendering function is defined, as introduced by \cite{ye2023gaussian}, allowing the splatting of Gaussian features $\mathbf{f}_i^k \in \mathbb{R}^D$, where $\mathbf{f}_i^k$ are the features attached to $\mathbf{g}^{k}_i$, to the feature image $\mathbf{Z}_j^k \in \mathbb{R}^{H \times W \times D}$. This feature rendering function can be expressed mathematically by:
\begin{equation}
\mathbf{Z}_j^k(u,v) = \sum_{i \in N'} \mathbf{f}_i^k \alpha_i^k \prod_{m=1}^{i-1} (1 - \alpha_m^k).
\label{eq:featureblending}
\end{equation}

\section{Proposed method}

\begin{figure}
    \centering
    \includegraphics[width=0.8\linewidth]{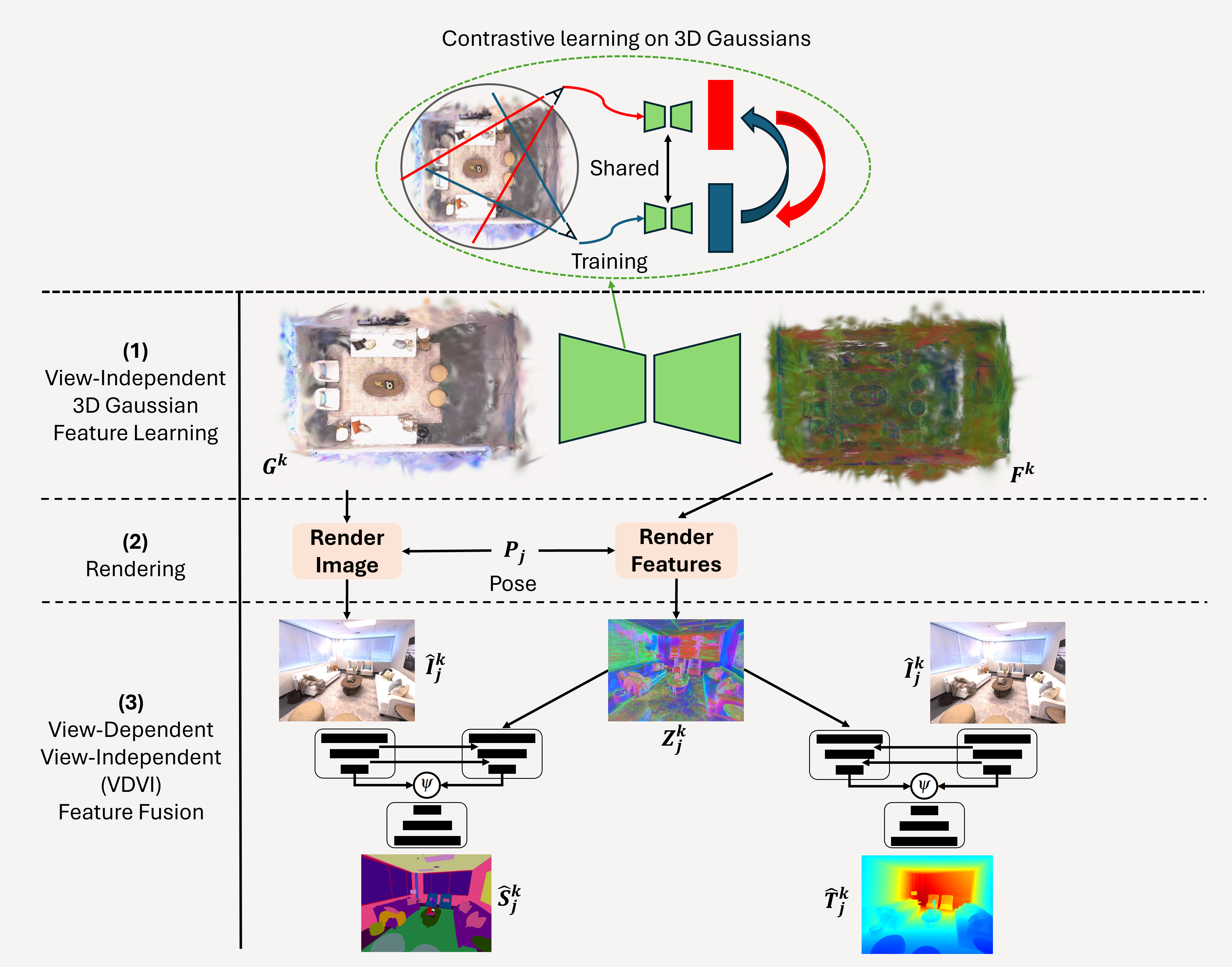}
    \caption{Overview of the proposed method.}
    \label{fig:arch}
\end{figure}

\subsection{Overview of the proposed method}






The architecture of the proposed method is illustrated in Figure \ref{fig:arch} and consists out of three main stages: (i) a new self-supervised view-independent 3D Gaussian feature extractor, (ii) the rendering of the 3D information encapsulated in the enhanced 3D Gaussians to a specific view, and (iii) a novel View-Dependent / View-Independent (VDVI) feature fusion. The first stage, described in Section \ref{subsec:prop_VI3D}, transforms a 3D Gaussian representation $\mathbf{G}^{k}$ of a scene $k$, into a set of features 
$\mathbf{F}^{k}= \{\mathbf{f}^{k}_{1}, \mathbf{f}^{k}_{2}, \ldots, \mathbf{f}^{k}_{N}\} \in \mathbb{R}^{N \times D}$, where $\mathbf{f}^{k}_{i}$ are the features learned for the corresponding 3D Gaussian $\mathbf{g}^k_{i}$, and $N$ is the number of 3D Gaussians in $\mathbf{G}^{k}$. Important to note is that, since the proposed method operates on the entire 3D Gaussian representation as input, the features $\mathbf{f}^{k}_{i}$ 
are view-independent. In the second stage, $\mathbf{F}^{k}$ is rendered by using alpha blending, described in Equation \ref{eq:featureblending}, to a specific view $j$, defined by the given pose $\mathbf{p}_j$. The feature image obtained by the splatting of $\mathbf{F}^{k}$ is denoted by $\mathbf{Z}^k_j \in \mathbb{R}^{H \times W \times D}$. In parallel, the novel view $\hat{\mathbf{I}}^k_j$ is rendered. In the last stage, the novel VDVI feature fusion module, described in Section \ref{subsec:prop_fusion}, extracts view-dependent features from $\hat{\mathbf{I}}^k_j$ and fuses them at different scales with $\mathbf{Z}^k_j$. A joint decoder is employed to obtain the semantic predictions $\hat{\mathbf{S}}^k_j$. By training the model on different scenes $k$, RT-GS2 is able to generalize semantic segmentation to unseen scenes.

\subsection{View-independent 3D Gaussian feature learning}
\label{subsec:prop_VI3D}




%


In order to obtain view-independent 3D Gaussian features, suitable for generalization on unseen scenes, we propose to learn 3D features directly from the Gaussian 3D representation. This is in stark contrast to existing methods~\cite{ye2023gaussian, dou2024cosseggaussians} which employ loss-terms in 2D to learn per-Gaussian features, leading to view-specific and most importantly, scene-specific Gaussian features. Our fundamentally different approach exploits the spatial distribution of the 3D Gaussians and inter-Gaussian relations to extract 3D features for unseen scenes. More specifically, we employ a point cloud autoencoder as backbone as it allows to process unstructured 3D points and retrieve global and local features. In this autoencoder, each 3D Gaussian is represented by its location while the additional properties are encoded in the channel dimension. Important to note, is that the 3D Gaussian feature extractor operates on the complete Gaussian 3D representation during inference. This ensures the retrieval of view-independent 3D Gaussian features, exploiting information from the entire scene, beyond a specific view. From Figure \ref{fig:3D_feat} can be seen that the obtained 3D features are view-consistent and this, by consequence, improves view-consistency of the final semantic segmentation $\hat{\mathbf{S}}^k_j$. An additional advantage of processing the entire scene is the possibility to compute the 3D features only once for the whole scene, further reducing required online inference time while navigating through the scene.

Lastly, we have opted to train the view-independent 3D Gaussian feature extractor in a self-supervised manner, following the contrastive learning training procedure of~\cite{xie2020pointcontrast}. This not only allows to perform the feature computation entirely in the 3D domain, but also ensures the extraction of robust features, suitable for multiple downstream tasks. In the supplementary, we ablate the robustness of the features by performing depth prediction. The employed self-supervised loss is described in Section \ref{subsec:loss}.

\subsection{View-Dependent / View-Independent (VDVI) feature fusion}
\label{subsec:prop_fusion}

The View-Dependent / View-Independent (VDVI) feature fusion module consists out of two parallel encoders: $Enc_{VD}$ and $Enc_{VI}$. The former encodes the image $\hat{\mathbf{I}}^k_j$ to view-dependent features. The latter takes the splatted view-independent 3D features, $\mathbf{Z}^k_j$, as input. At different scales, the view-independent features are fused with the encoded view-dependent features in encoder $Enc_{VI}$. At the lowest scale, the output of both encoders are fused by a fusion function $\psi$. Hereafter, the resulting features are employed by a decoder $Dec$ to retrieve the semantic predictions $\hat{\mathbf{S}}^k_j$. This can be expressed mathematically as follows:
\begin{equation}
\hat{\mathbf{S}}^k_j = Dec(\psi(Enc_{VD}(\hat{\mathbf{I}}^k_j), Enc_{VI}(\mathbf{Z}^k_j, \hat{\mathbf{I}}^k_j))).
\label{eq:vdvi_fusion}
\end{equation}
VDVI feature fusion improves pure segmentation performance and increases view-consistency of the results.

\subsection{Loss}
\label{subsec:loss}
The self-supervised 3D Gaussian feature extractor and VDVI feature fusion are optimized with respect to \( L_{cl} \)~\cite{xie2020pointcontrast} and $\mathcal{L}_{sem}$, respectively, each one defined below. Detailed descriptions on the losses and individual loss-terms can be found in the supplementary material.

\begin{equation}
L_{cl} = -\sum_{(m,n) \in P_k} \log \frac{\exp(f^k_m \cdot f^k_n / \tau)}{\sum_{(l,\cdot) \in P_k} \exp(f^k_m \cdot f^k_l / \tau)}
\label{eq:loss_CL}
\end{equation}

\begin{equation}
\mathcal{L}_{sem} = \mathcal{L}_{CrossEntropy} + \lambda_{CeCo} \mathcal{L}_{CeCo}.
\label{eq:total_loss}
\end{equation}

\section{Experiments}

\subsection{Experimental setup}

\textbf{Datasets and Metrics.} 
Experiments were conducted on three datasets: Replica~\cite{straub2019replica}, ScanNet~\cite{dai2017scannet}, and ScanNet++~\cite{yeshwanth2023scannet++}, representing synthetic and real-world indoor scenes. Experimental settings meticulously followed those of \cite{chou2024gsnerf} for Replica and ScanNet. Details and splits are provided in the supplementary material. As evaluation metrics, we employed mean Intersection over Union (mIoU), mean Accuracy (mAcc), and overall Accuracy (oAcc) for segmentation, and Peak Signal-to-Noise Ratio (PSNR), Structural Similarity (SSIM), and Learned Perceptual Image Patch Similarity (LPIPS) for rendering quality. Inference speed was measured in Frames Per Second (FPS).

\noindent \textbf{Implementation details.}
To train the RT-GS2 model, RGB data and their corresponding semantic masks are required for the training scenes. During testing, the model is able to operate solely on the Gaussian Splatting representation, discarding the need for expensive semantic masks for unseen scenes. Additionally to generalization performance, we also report the performance of our model after fine-tuning the generalized model for a fixed number of iterations using the scene's semantic labels, allowing an increased performance.
The view-independent 3D Gaussian feature extractor employs a PointTransformerV3~\cite{wu2023point} with the output feature dimension $D$ empirically chosen as 32. The input, with channel dimension 10 (xyz, base color, scale information and opacity), is subsampled during training with voxelization (size 0.07). The contrastive loss is computed among 4096 corresponding points from 2 different views. Asymformer~\cite{du2023asymformer} is selected as a real-time VDVI feature fusion backbone. $\lambda_{CeCo}$ is chosen 0.4 and LSR~\cite{szegedy2016rethinking} is employed for generalization. All experiments were performed using a NVIDIA GeForce RTX 3090 GPU.

\subsection{Comparison to state of the art}


\begin{table}
    \small
    \centering
    \tabcolsep=0.11cm
    \begin{tabular}{|c|c|c|c|c|c||c|c|c||c|} 
        \cline{4-10}
        \multicolumn{3}{c|}{} & \multicolumn{3}{c||}{Semantic} & \multicolumn{3}{c||}{Rendering} & \multicolumn{1}{c|}{Time}\\
        \cline{2-10}
        \multicolumn{1}{c|}{} & Method & Published & mIoU & mAcc & oAcc & PSNR$\uparrow$ & SSIM$\uparrow$ & LPIPS$\downarrow$ & FPS$\uparrow$\\ 
        \hline
        \multirow{5}{*}{General.} & MVSNeRF*~\cite{johari2022geonerf} & CVPR2022 & 30.21 & 39.75 & 69.35 & 23.68 & 84.37 & 28.08 & -\\
        & Neuray*~\cite{liu2022neural} & CVPR2022 & 40.91 & 50.15 & 76.23 & 27.80 & 89.55 & 23.68 & $<$0.03\\
        & S-Ray~\cite{liu2023semantic} & CVPR2023 & 43.27 & 52.85 & 77.63 & 26.77 & 88.54 & 22.81 & 0.03\\ 
        & GSNeRF~\cite{chou2024gsnerf} & CVPR2024 & 51.23 & 61.10 & 83.06 & 31.71 & 92.89 & 12.93 & -\\
        \cline{2-10}        & Ours & - & \bf{59.24} & \bf{66.04} & \bf{93.99} & \bf{36.02} & \bf{97.12} & \bf{4.91}& \bf{27.03}\\ 
        \hline
        \hline        \multirow{2}{*}{Finetune} & S-Ray~\cite{liu2023semantic} & CVPR2023 & 84.12 & 88.53 & 96.36 & 27.78 & 84.53 & 12.88 & 0.03\\
        \cline{2-10}
        & Ours & - & \bf{93.75} & \bf{96.19} & \bf{99.33} & \bf{36.02} & \bf{97.12} & \bf{4.91} & \bf{27.03}\\ 
        \hline
    \end{tabular}
    \caption{Comparison on REPLICA~\cite{straub2019replica} of the proposed method against the state of the art for generalizable semantic segmentation and after finetuning on a specific scene. * denotes the addition of a semantic head.}
    \label{tab:replica_results}
\end{table}
\begin{table}
    \small
    \centering
    \tabcolsep=0.11cm
    \begin{tabular}{|c|c|c|c|c|c||c|c|c||c|} 
        \cline{4-10}
        \multicolumn{3}{c|}{} & \multicolumn{3}{c||}{Semantic} & \multicolumn{3}{c||}{Rendering} & \multicolumn{1}{c|}{Time}\\
        \cline{2-10}
        \multicolumn{1}{c|}{} & Method & Published & mIoU & mAcc & oAcc & PSNR$\uparrow$ & SSIM$\uparrow$ & LPIPS$\downarrow$ & FPS$\uparrow$\\ 
        \hline
        \multirow{5}{*}{General.} & MVSNeRF*~\cite{johari2022geonerf} & CVPR2022 & 43.06 & 53.63 & 66.90 & 24.14 & 80.36 & 34.63 & -\\
        & Neuray*~\cite{liu2022neural} & CVPR2022 & 46.09 & 53.79 & 66.39 & 25.24 & 84.39 & 31.33 & $<$0.11\\
        & S-Ray~\cite{liu2023semantic} & CVPR2023 & 47.69 & 54.47 & 64.90 & 25.13 & 84.18 & 30.44 & 0.11\\
        & GSNeRF~\cite{chou2024gsnerf} & CVPR2024 & 52.21 & 60.14 & 74.71 & \bf{31.49} & \bf{90.39} & \bf{13.87} & -\\
        \cline{2-10}
        & Ours & - & \bf{53.27} & \bf{62.43} & \bf{81.20} & 27.27 & 89.10 & 21.77 & \bf{27.03}\\
        \hline
        \hline
        \multirow{3}{*}{Finetune} & S-Ray~\cite{liu2023semantic} & CVPR2023 & 91.6 & 92.2 & 97.3 & 27.31 & - & - & 0.11\\
        & GSNeRF~\cite{chou2024gsnerf} & CVPR2024 & 93.2 & 96.8 & 98.2 & \bf{30.89} & - & - & -\\
        \cline{2-10}
        & Ours & - & \bf{96.94} & \bf{98.87} & \bf{99.01} & 27.27 & 89.10 & 21.77 & \bf{27.03}\\
        \hline
    \end{tabular}
    \caption{Comparison on ScanNet~\cite{dai2017scannet} of the proposed method against the state of the art for generalizable semantic segmentation and after finetuning on a specific scene. * denotes the addition of a semantic head.}
    \label{tab:scannet_results}
\end{table}

\begin{figure}%
    \centering
    \includegraphics[width=1\linewidth]{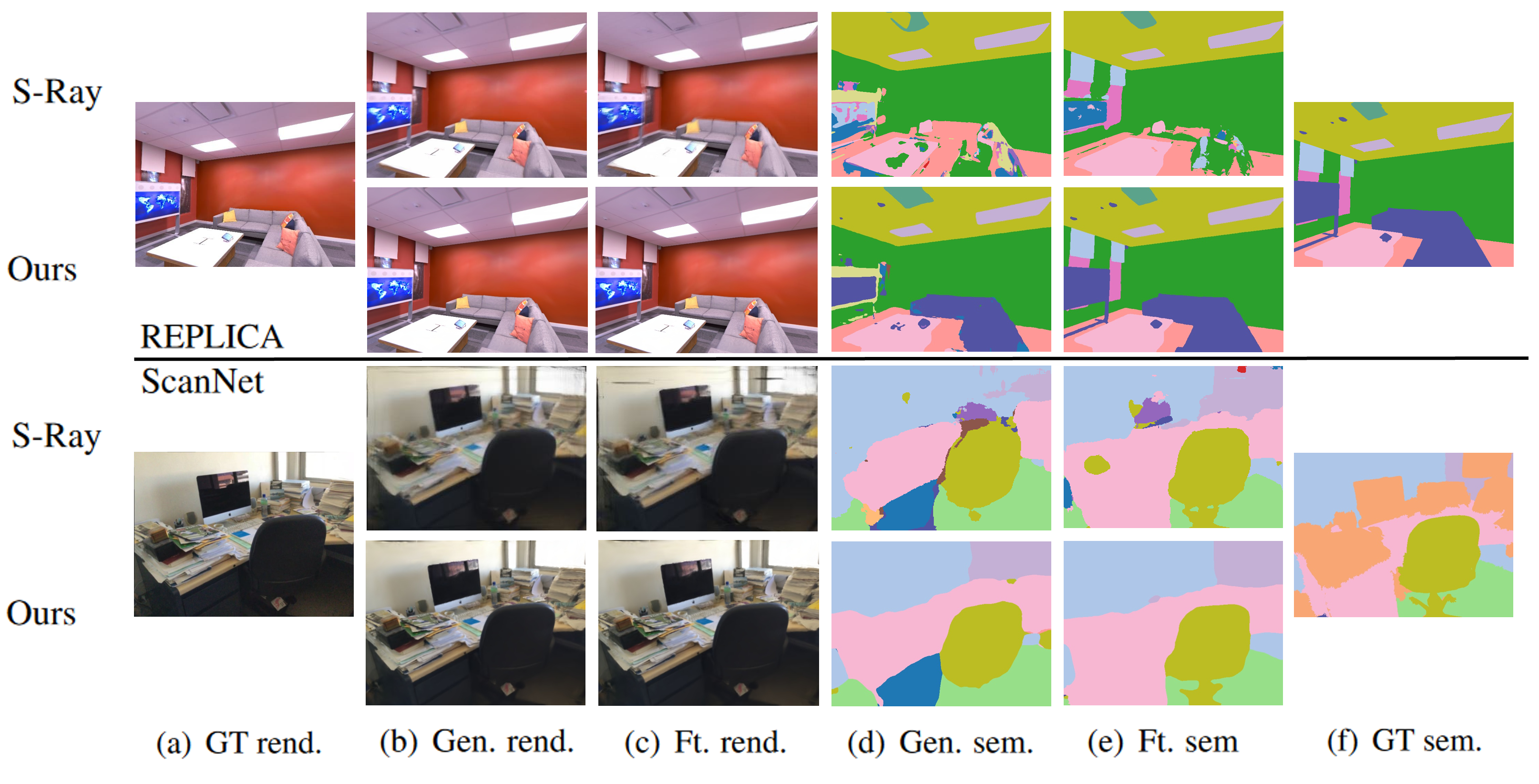}
    \caption{Qualitative results on Replica and ScanNet. The table presents generalizable (gen.) and finetuning (ft.) results on Replica (first two rows) and ScanNet (last two rows) for both rendering (rend.) and semantic segmentation (sem.). Comparisons between RT-GS2 and Semantic-Ray\cite{liu2023semantic} are made.}%
    \label{fig:qualitative_results}%
\end{figure}

\noindent \textbf{Semantic segmentation on Replica.} Table \ref{tab:replica_results} showcases our results on Replica~\cite{straub2019replica} alongside a comparison with state-of-the-art methods. Our method significantly outperforms existing approaches in both segmentation quality and inference time. Specifically, for semantic generalization, RT-GS2 outperforms the state of the art across all evaluated metrics, achieving an impressive 8.01\% increase in mIoU. Fine-tuning on specific scenes for 20k iterations further improves performance to 93.75\% and 99.33\% for mIoU and oAcc, respectively. While RT-GS2 does not present rendering generalization, we also present rendering performance for completeness. The usage of Gaussian Splatting for rendering allows for an increase in rendering performance. Notably, our method achieves real-time novel view synthesis and segmentation at 27.03 FPS, a remarkable 901 times speedup compared to S-Ray. Qualitative results, depicted in Figure \ref{fig:qualitative_results}, demonstrate compelling segmentation performance. After finetuning, even the fine details are correctly segmented. RT-GS2 consistently outperforms S-Ray in both settings. Additional visual results and a video can be found in the supplementary material.

\noindent \textbf{Semantic segmentation on ScanNet.} 
Table \ref{tab:scannet_results} displays our results on the real-world dataset ScanNet~\cite{dai2017scannet}. While Gaussian Splatting does not surpass NeRF-based GSNeRF~\cite{chou2024gsnerf} in rendering quality on ScanNet, likely due to the high presence of motion blur and other sources of noise, our proposed method is still capable of consistently outperforming existing methods in segmentation quality and inference time, both for generalization and finetuning (5k iterations). The qualitative results in Figure \ref{fig:qualitative_results} confirm these observations.


\noindent \textbf{Semantic segmentation on ScanNet++.} Additionally, we present results on ScanNet++\cite{yeshwanth2023scannet++}. While this dataset was not publicly available during the publication of prior works \cite{johari2022geonerf, liu2022neural, liu2023semantic, chou2024gsnerf}, ScanNet++ is well-suited for novel view synthesis and generalizable semantic segmentation, exhibiting high-quality images and accurately annotated classes. In Table \ref{tab:scannetpp_results}, the proposed method demonstrates solid performance, enabling future comparisons. Qualitative results can be found in the supplementary material.


\begin{table}
    \small
    \centering
    \tabcolsep=0.11cm
    \begin{tabular}{|c|c|c|c|c||c|c|c||c|} 
        \cline{3-9}
        \multicolumn{2}{c|}{} & \multicolumn{3}{c||}{Semantic} & \multicolumn{3}{c||}{Rendering} & \multicolumn{1}{c|}{Time}\\
        \cline{2-9}
        \multicolumn{1}{c|}{} & Method & mIoU & mAcc & oAcc & PSNR$\uparrow$ & SSIM$\uparrow$ & LPIPS$\downarrow$ & FPS$\uparrow$\\  
        \hline
        \multirow{1}{*}{Generalization} & Ours & \bf{66.14} & \bf{77.32} & \bf{83.79} & \bf{26.39} & \bf{87.31} & \bf{17.08} & \bf{27.03}\\
        \hline
        \hline
        \multirow{1}{*}{Finetuned} & Ours & \bf{91.85} & \bf{95.73} & \bf{96.24} & \bf{26.39} & \bf{87.31} & \bf{17.08} & \bf{27.03}\\
        \hline
    \end{tabular}
    \caption{Quantitative results on ScanNet++~\cite{yeshwanth2023scannet++} of the proposed method for generalizable semantic segmentation and after finetuning on a specific scene.}
    \label{tab:scannetpp_results}
\end{table}

\subsection{Ablation study}


\noindent \textbf{Robustness of self-supervised features.} In this study, we investigate the robustness of the self-supervised view-independent 3D Gaussian features for a different task: depth prediction. Quantitative and qualitative results are available in the supplementary material.



\noindent \textbf{Evaluation of view-dependent and view-independent features.} 
Table \ref{tab:ablation} quantifies the influence of the synthesized image $\hat{\mathbf{I}}_j^k$ and rendered view-independent 3D Gaussian features $\mathbf{Z}_j^k$. Removal of $\hat{\mathbf{I}}_j^k$ and $\mathbf{Z}_j^k$ results in significant decreases in mAcc by 22.17\% and 4.48\%, respectively. These findings highlight the vital importance of $\hat{\mathbf{I}}_j^k$ for semantic segmentation performance. Additionally, the inclusion of view-independent features not only enhances performance but also improves view-consistency, as demonstrated in Figure \ref{fig:3D_feat}.



\noindent \textbf{Evaluation of loss-terms.} 
In Table \ref{tab:ablation}, we examine the impact of the label smoothing regularizer (LSR)~\cite{szegedy2016rethinking} and loss-term $\mathcal{L}_{CeCo}$~\cite{zhong2023understanding} through ablation. Results show that both components contribute positively to generalization performance. While the LSR does not provide benefits for the top20 frequent classes when finetuned, it increases performance when all classes are taking into account, suggesting its importance for the less frequent classes.


\begin{table}
    \small
    \centering
    \tabcolsep=0.11cm
    \begin{tabular}{|c|c|c|c|c||c|c|c|c|c|c|} 
        \cline{2-11}
        \multicolumn{1}{c|}{} & \multicolumn{2}{c|}{VDVI features} & \multicolumn{2}{c||}{Loss} & \multicolumn{3}{c|}{Semantic - top20} & \multicolumn{3}{c|}{Semantic - all}\\
        \cline{2-11}
        \multicolumn{1}{c|}{} & \phantom{ee}$\hat{\mathbf{I}}_j^k$\phantom{ee} & $\mathbf{Z}_j^k$ & \phantom{  }LSR\phantom{  } 
 & $\mathcal{L}_{CeCo}$ & mIoU & mAcc & oAcc & mIoU & mAcc & oAcc\\
        \hline
        \multirow{5}{*}{Generalization} & \xmark & \cmark & \cmark & \cmark & 37.43 & 43.87 & 87.26 & 27.05 & 31.49 & 83.02\\
        & \cmark & \xmark & \cmark & \cmark & 55.57 & 61.56 & 93.51 & 44.41 & 51.38 & 89.71\\
        & \cmark & \cmark & \xmark & \cmark & 58.6 & 65.07 & 93.92 & 45.90 & 54.30 & 90.13\\ 
        & \cmark & \cmark & \cmark & \xmark & 58.77 & 65.38 & \bf{94.04} & 47.11 & 55.59 & 90.51\\
        \cline{2-11}
        & \cmark & \cmark & \cmark & \cmark & \bf{59.24} & \bf{66.04} & 93.99 & \bf{49.11} & \bf{57.04} & \bf{90.68}\\
        \hline
        \hline
        \multirow{3}{*}{Finetuned} & \cmark & \cmark & \xmark & \xmark & 93.11 & 96.15 & 99.16 & 86.05 & 91.17 & 99.07\\
        & \cmark & \cmark & \cmark & \cmark & 92.11 & 94.81 & 99.25 & \bf{92.48} & \bf{94.80} & \bf{99.25}\\
        \cline{2-11}
        & \cmark & \cmark & \xmark & \cmark & \bf{94.31} & \bf{96.19} & \bf{99.33} & 89.34 & 93.13 & 99.22\\
        \hline
    \end{tabular}
    \caption{Ablation study on Replica~\cite{straub2019replica} for both the 20 most frequent and all classes.}
    \label{tab:ablation}
\end{table}


\section{Limitations}


Despite the significant improvements in both performance and speed achieved by RT-GS2, some limitations exist. While RT-GS2 enhances view-consistency for semantic segmentation by learning and leveraging view-independent 3D features (as qualitatively demonstrated in Figure \ref{fig:3D_feat}), this approach does not fully guarantee strict view consistency across all perspectives. More specifically, while the primary elements of the scene generally remain stable, minor flickering can occur in smaller regions when the viewpoint changes.

\section{Conclusion}

This paper presents a novel real-time generalizable semantic segmentation method employing Gaussian Splatting. Through extensive experimentation, we have demonstrated its superiority over existing methods in both semantic segmentation quality and real-time performance, achieving significant improvements in mIoU on the Replica dataset and a remarkable 901 times speedup compared to current approaches. RT-GS2 represents a significant advancement in the field, providing the first real-time generalizable semantic segmentation method for 3D Gaussian representations of radiance fields.

\section*{Acknowledgement}

This work is funded by Innoviris within the research project SPECTRE and by Research Foundation Flanders (FWO) within the research project G094122N.

\bibliography{egbib}

\newpage
\appendix
\section{Robustness of self-supervised features.}

In order to evaluate the robustness of the self-supervised view-independent 3D Gaussian features we utilize them for another downstream task, namely depth prediction. Instead of doing monocular depth prediction on the rendered image, we employ our splatted 3D Gaussian features as additional information. While we did not re-train the 3D Gaussian feature learning, we trained another VDVI feature fusion module and depth prediction head. A schematic overview of the employed architecture can be found in Figure 2 of the main body. To evaluate our results quantitatively, we employed the popular metrics, employed in \cite{murez2020atlas}. In Table \ref{tab:depth_metrics}, we describe the different employed metrics, where $d$ and $d^*$ are the true and predicted depth values of a pixel, respectively, and $n$ the total number of pixels in the instance. To train the VDVI feature fusion module and depth prediction head, we employed the MSE-loss as loss function. 

\begin{figure}
\centering
\begin{tabular}{>{\centering\arraybackslash}m{3.2cm}>{\centering\arraybackslash}m{3.2cm}>{\centering\arraybackslash}m{3.2cm}}
\includegraphics[width=3.5cm]{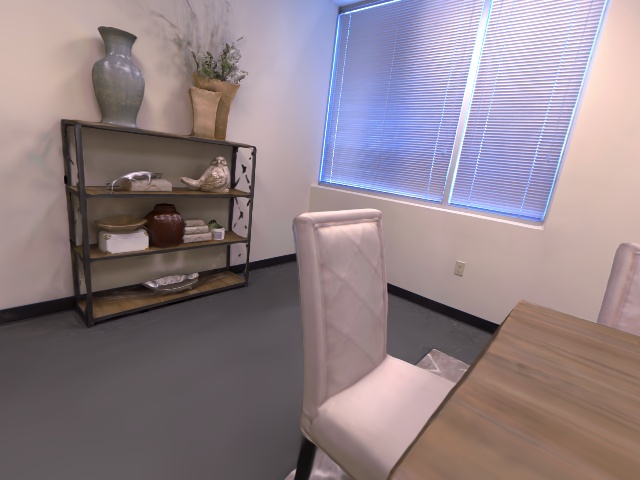} & \includegraphics[width=3.5cm]{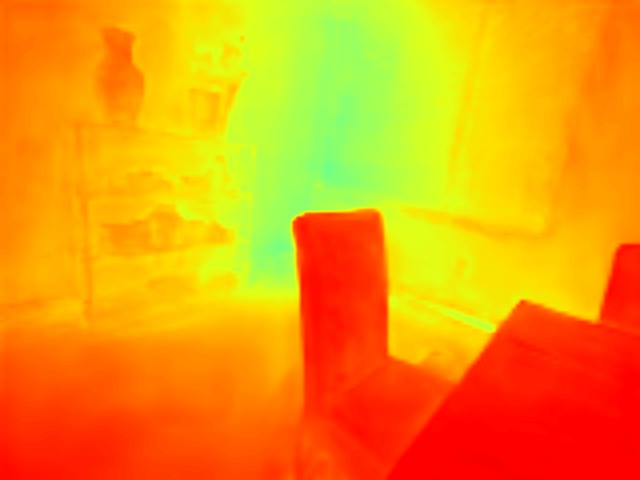} & \includegraphics[width=3.5cm]{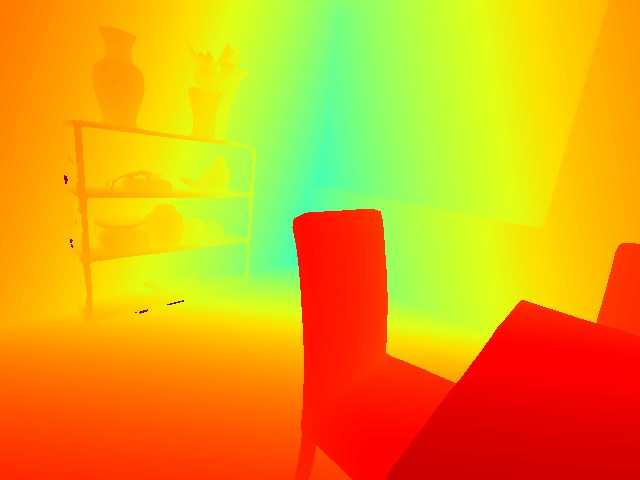}\\
\includegraphics[width=3.5cm]{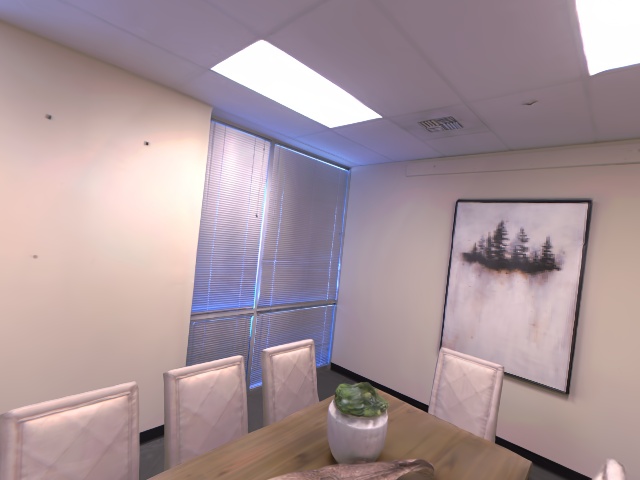} & \includegraphics[width=3.5cm]{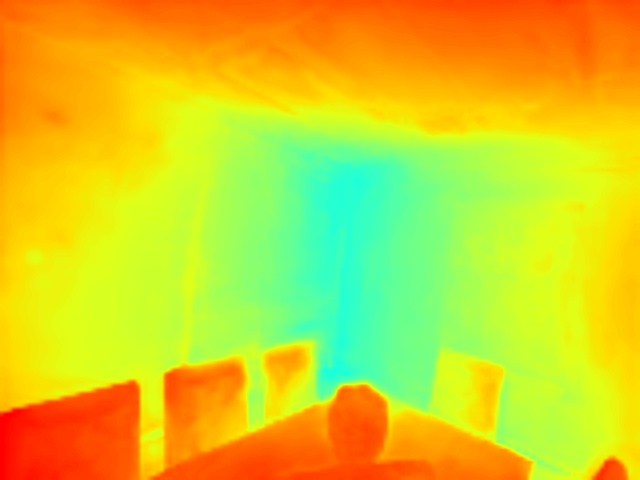} & \includegraphics[width=3.5cm]{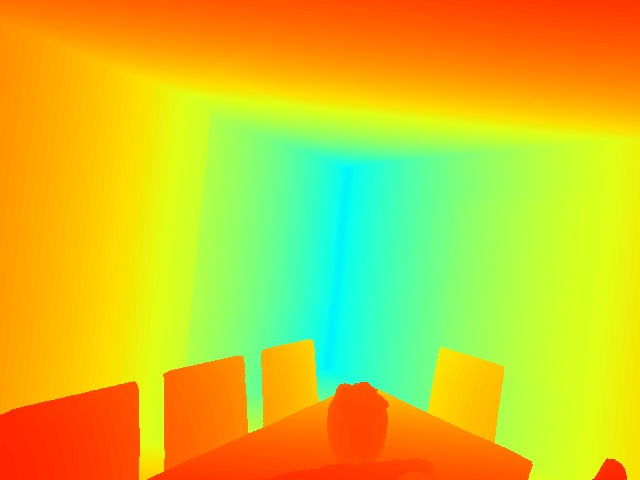} \\
\includegraphics[width=3.5cm]{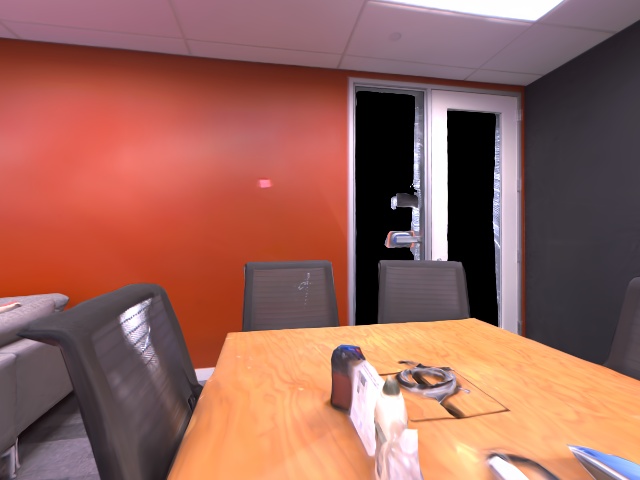} & \includegraphics[width=3.5cm]{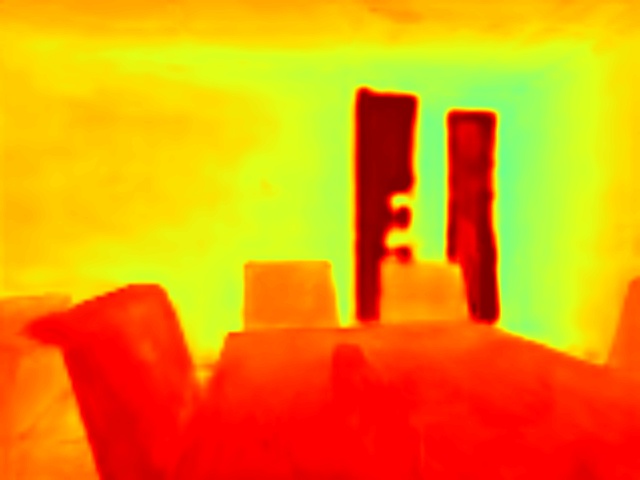} & \includegraphics[width=3.5cm]{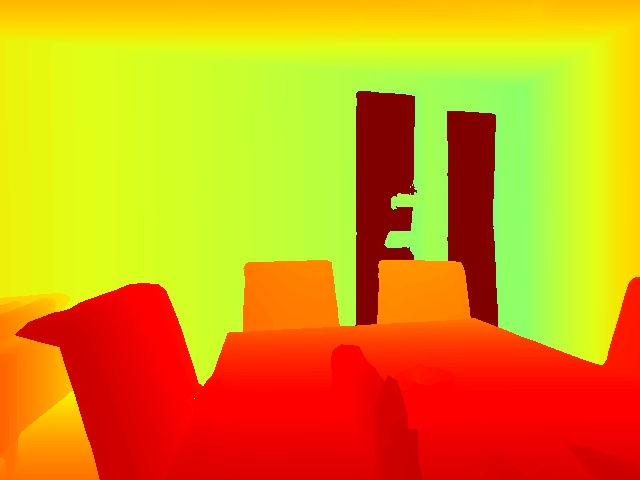}\\
\includegraphics[width=3.5cm]{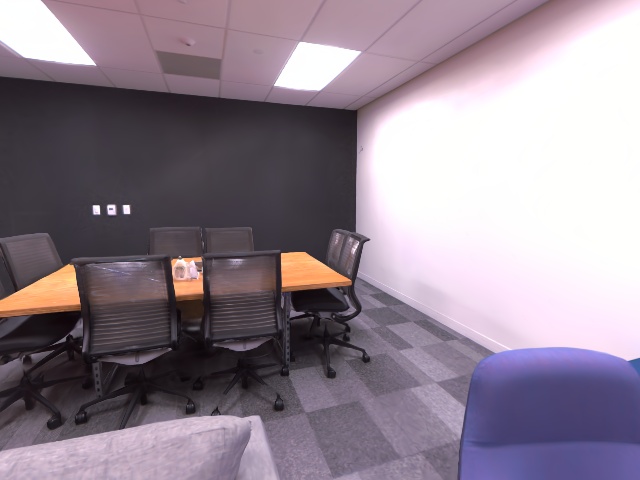} & \includegraphics[width=3.5cm]{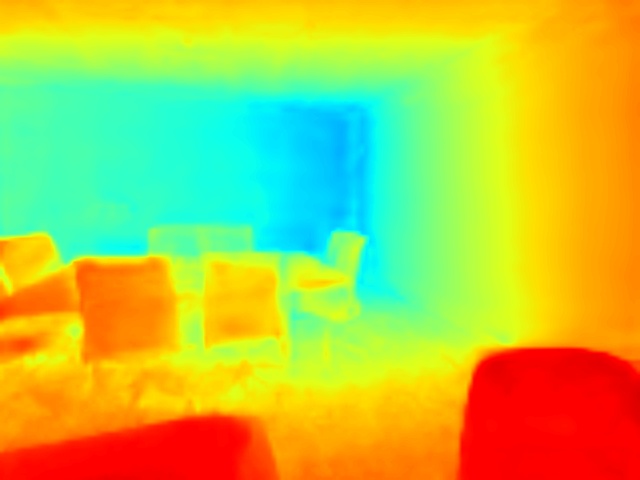} & \includegraphics[width=3.5cm]{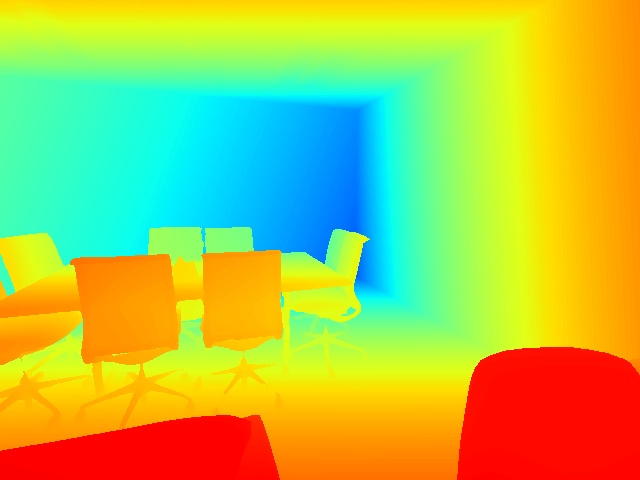}\\
\includegraphics[width=3.5cm]{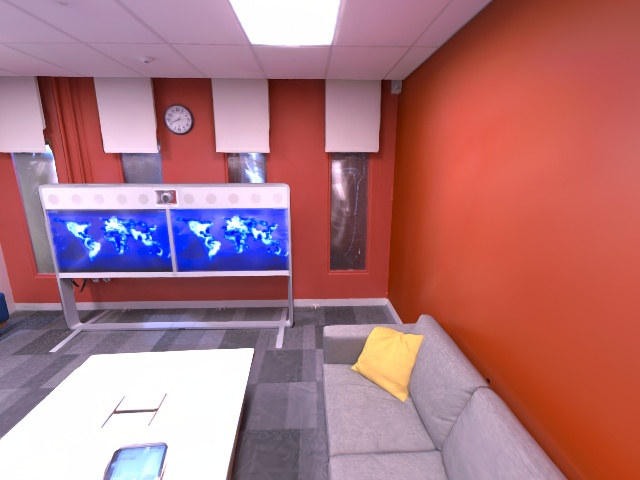} & \includegraphics[width=3.5cm]{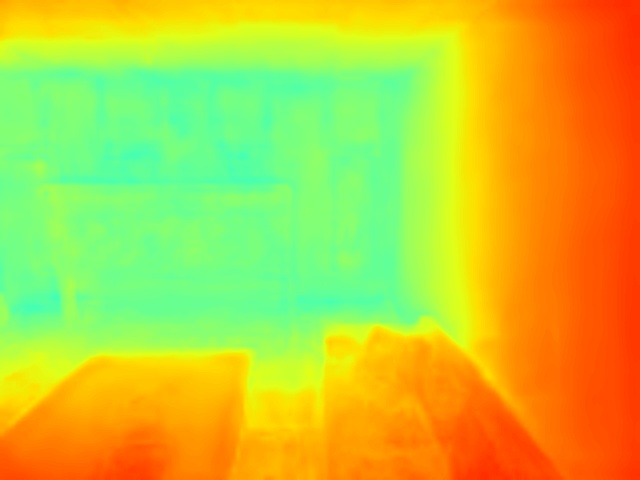} & \includegraphics[width=3.5cm]{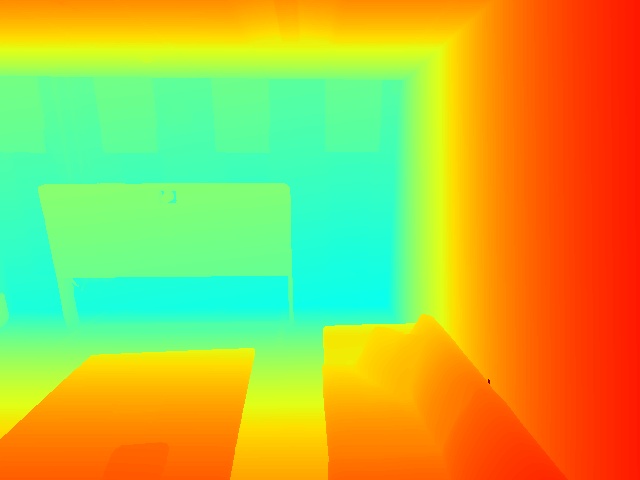}\\
Rendering & Depth pred. & Depth GT \\
\end{tabular}
\caption{Depth prediction of the proposed method on Replica dataset.}
\label{tab:depth_qual}
\end{figure}

The results of the proposed method and ablation of the 3D Gaussian features on the Replica dataset~\cite{straub2019replica} can be found in Table \ref{tab:depth_predictions}. It can be seen that the addition of our view-independent 3D Gaussian features, allows a consistent improvement of the depth prediction for all employed metrics. More specifically, we achieve an important 24.1\% improvement in Abs. Rel. and 25.4\% in RMSE, compared to the experiment without self-supervised view-independent 3D Gaussian features, i.e. monocular depth prediction. Qualitative results are presented in Figure \ref{tab:depth_qual}, visualized using a heatmap going from red, closeby, to blue, far away. It can be noticed that the depth predictions are of high quality, both closeby and far away, closely resembling the ground-truth depth maps.

\begin{table}[h!]
\centering
\begin{tabular}{|c|c|}
\hline
Abs Rel & $\frac{1}{n} \sum \frac{|d - d^*|}{d^*}$ \\ \hline
Abs Diff & $\frac{1}{n} \sum |d - d^*|$ \\ \hline
Sq Rel & $\frac{1}{n} \sum \frac{|d - d^*|^2}{d^*}$ \\ \hline
RMSE & $\sqrt{\frac{1}{n} \sum |d - d^*|^2}$ \\ \hline
$\delta < 1.25^i$ & $\frac{1}{n} \sum \left( \max \left( \frac{d}{d^*}, \frac{d^*}{d} \right) < 1.25^i \right)$ \\ \hline
Comp & \% valid predictions \\ \hline
\end{tabular}
\caption{Definition of the employed depth metrics~\cite{murez2020atlas}}
\label{tab:depth_metrics}
\end{table}



\begin{table}[h!]
\centering
\begin{tabularx}{\textwidth}{|c|X|X|X|X|X|X|X|X|}
\hline
\small & \small Abs Rel & \small Abs Diff & \small Sq Rel & \small RMSE & \small $\delta$<1.25 & \small $\delta$<$1.25^2$ & \small $\delta$<$1.25^3$ & \small Comp\\ \hline
Ours w/o 3D features & 0.108 & 0.238 & 0.040 & 0.303 & 0.911 & 0.985 & 0.994 & 0.998\\ 
\hline
Ours & \bf{0.082} & \bf{0.173} & \bf{0.024} & \bf{0.226} & \bf{0.948} & \bf{0.994} & \bf{0.998} & \bf{0.999}\\ 
\hline
\end{tabularx}
\caption{Performance metrics on Replica dataset~\cite{straub2019replica} and ablation of the effect of the self-supervised view-independent 3D Gaussian features for depth prediction. W/o stands for without.}
\label{tab:depth_predictions}
\end{table}

\section{Details of the loss functions}

\subsection{Contrastive loss}

Our robust view-independent 3D features are learnt in a self-supervised manner, using the contrastive point cloud loss described in \cite{xie2020pointcontrast}. Additionaly, specifically for operating on 3D Gaussian representations, we added scale and opacity transformations to the training input to ensure that the existing information of the Gaussians is retained during training. We will follow below the notations from the main body of our paper, for consistency reasons and facilitate reader comprehension.


Given a set of \(k\) scenes with their corresponding \(\mathbf{G}^{k}\) representation, we construct a list of correspondences \(P_{1}, \ldots, P_{k}\). For two different viewpoints of a scene \(k\), we define the set \(\{\mathbf{g}^{k}_{1}\prime, \mathbf{g}^{k}_{2}\prime, \ldots\}\) as the points that lie within the frustum of both views. Indices \(m\) and \(n\) denote the positions of the points in the first and second views, respectively, as listed in \(P_k\). These correspondences are considered positive pairs and retain their positive value for contrastive loss computation. The employed loss-function can be expressed as follows:
\begin{equation}
L_{cl} = -\sum_{(m,n) \in P_k} \log \frac{\exp(f^k_m \cdot f^k_n / \tau)}{\sum_{(l,\cdot) \in P_k} \exp(f^k_m \cdot f^k_l / \tau)},
\label{eq:loss_CL}
\end{equation}
where $\tau$ is a constant set to 0.07. 

\subsection{Semantic loss}

The employed semantic loss function is composed by two terms: the per pixel cross entropy loss and the $CeCo$ term, expressed mathematically as follows 
\begin{equation}
\mathcal{L}_{sem} = \mathcal{L}_{CrossEntropy} + \lambda_{CeCo} \mathcal{L}_{CeCo}.
\label{eq:total_loss}
\end{equation}
The $CeCo$ loss-term, $\mathcal{L}_{CeCo}$, described in \cite{zhong2023understanding}, can be expressed mathematically as follows:
\begin{equation}
\mathcal{L}_{CeCo}(\bar{Z}, \mathbf{W}^*) = - \sum_{k=1}^{K} \log \left( \frac{\exp (\bar{z}_k^\top \mathbf{w}_k^*)}{\sum_{k'=1}^{K} \exp (\bar{z}_{k'}^\top \mathbf{w}_{k'}^*)} \right),
\end{equation}
Where $\bar{Z}, \mathbf{W}^*$ are the features centers and the classifier weights, respectively. In this scenario $K$ describes the number of classes. The intuition of using this addition term was due to the highly unbalanced nature of the 2D segmentation labels, as classes such as walls, ceiling and floors are dominant. This intuition was supported by the results of our ablation study, in Table 4 of the main body, where can be seen that $CeCo$ proves mostly beneficial for the less dominant classes. This can be noticed as the performance increase for the experiment with all classes is larger than the performance increase for only 20 most frequent classes.
\section{Dataset setup}

In this section, we discuss our dataset setup in more detail. Our experiments were conducted on three different datasets: Replica \cite{straub2019replica}, ScanNet \cite{dai2017scannet}, and ScanNet++ \cite{yeshwanth2023scannet++}. For the Replica split, we followed the Semantic-NeRF setup as described in \cite{zhi2021place}. The tested resolution was \(480 \times 640\). For the ScanNet dataset, we trained on the first 60 scenes and tested on 10 scenes, following the setup in \cite{chou2024gsnerf}. The tested resolution was the same as in \cite{liu2023semantic, chou2024gsnerf}. For ScanNet++, we randomly selected 40 scenes for training and 10 scenes for testing.

\section{More implementation details}

\subsection{View-independent implementation details}

For the self-supervised view-independent 3D Gaussian feature learning, we used PointTransformerV3 \cite{wu2023point} as the encoder, optimized with the Adam optimizer~\cite{kingma2014adam} ($\beta_1 = 0.9$, $\beta_2 = 0.999$) and a weight decay of $10^{-5}$. The number of points queried for contrastive learning is 4096. To select the different views for scenes, we ensure that the corresponding frustums for those views have an overlap of at least 30\% but no more than 80\%.

\subsection{View-Dependent / View-Independent (VDVI) feature fusion implementation details}

We used the backbone of Asymformer \cite{du2023asymformer}, extracting the last activations before the final layer to provide the appropriate information for $L_{cl}$. Optimization was performed using the AdamW optimizer~\cite{loshchilov2017decoupled} ($\beta_1 = 0.9$, $\beta_2 = 0.999$) with a weight decay of $10^{-4}$. During the generalization stage, we applied a warm-up of 4 epochs, which was disregarded during the fine-tuning stage. The learning rate was set to $10^{-4}$.
 
\section{Video Demo}

Attached to the supplementary materials is also a video demo in which we exhibit the results for different video sequences, displaying the extraordinary performance of RT-GS2 over a complete sequence. Sequences were selected for both a synthetic dataset, Replica~\cite{straub2019replica}, and real-world data, ScanNet++~\cite{yeshwanth2023scannet++}, exhibiting the robustness of our highly accurate results. It can be noticed from the video demo, that the proposed method enhances view-consistency for our selected downstream tasks.


\section{Additional visualizations}

\begin{figure}[H]
\centering
\begin{tabular}{>{\centering\arraybackslash}m{2.7cm}>{\centering\arraybackslash}m{2.7cm}>{\centering\arraybackslash}m{2.7cm}>{\centering\arraybackslash}m{2.7cm}}
\includegraphics[width=3.0cm]{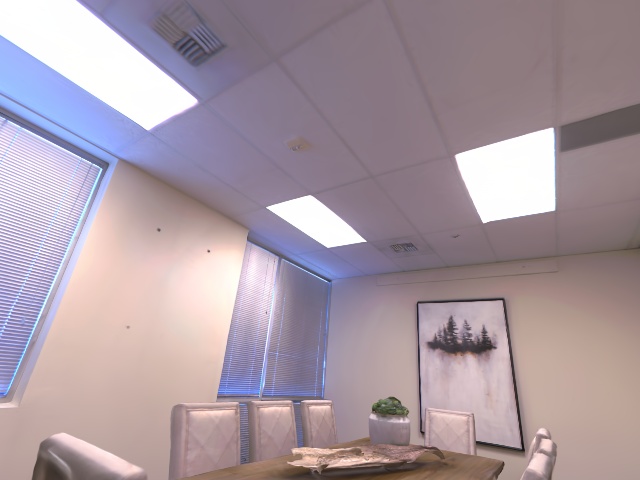} & \includegraphics[width=3.0cm]{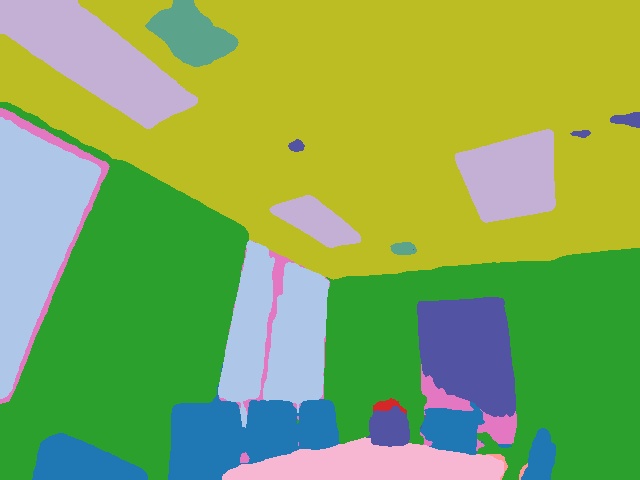} & \includegraphics[width=3.0cm]{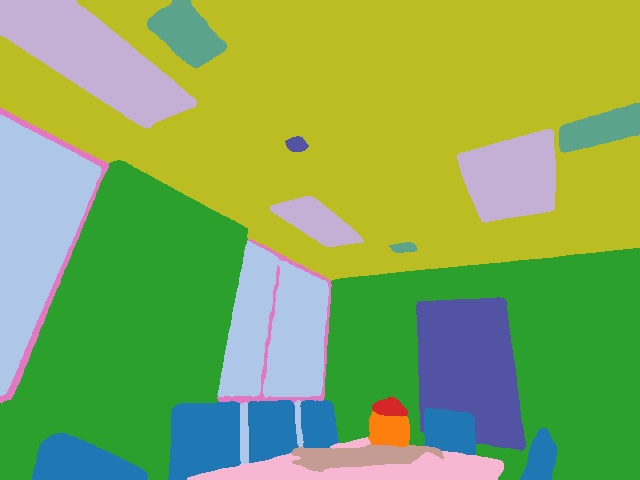} & \includegraphics[width=3.0cm]{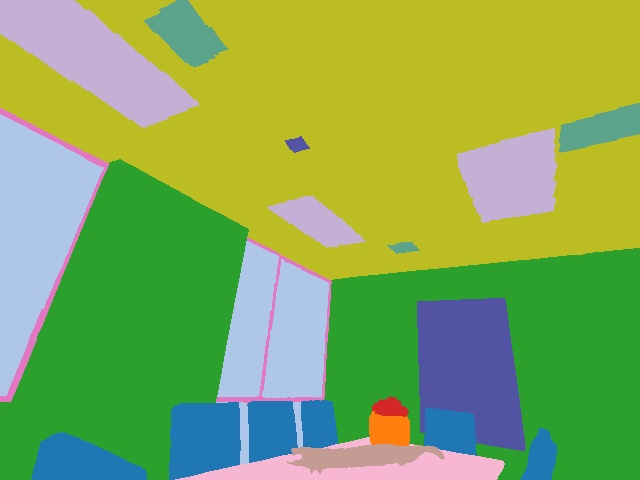} \\
\includegraphics[width=3.0cm]{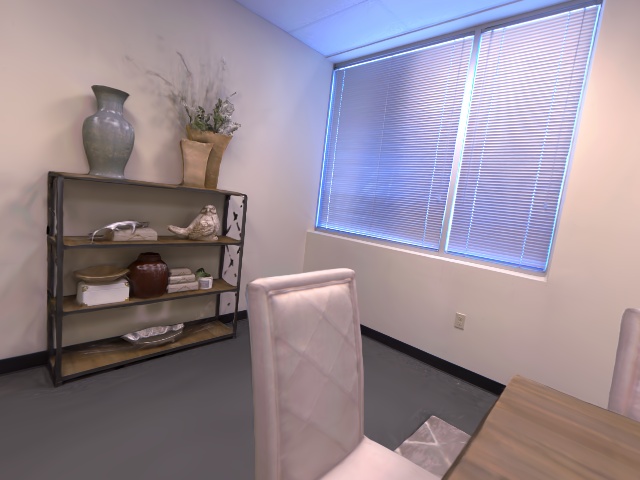} & \includegraphics[width=3.0cm]{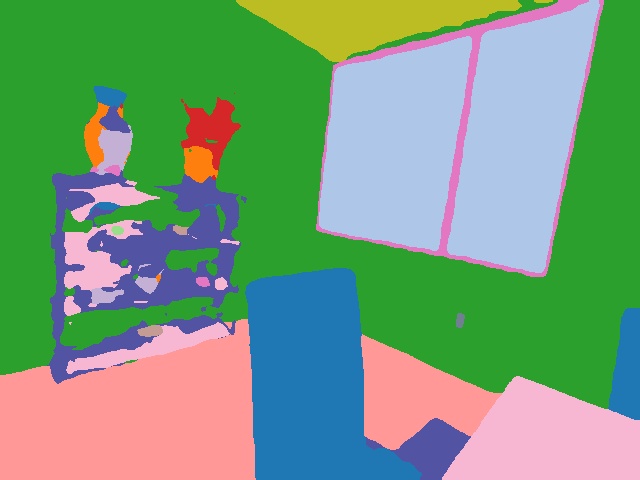} & \includegraphics[width=3.0cm]{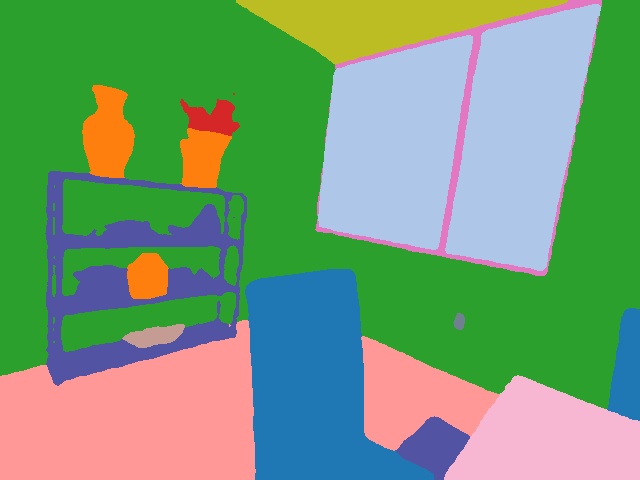} & \includegraphics[width=3.0cm]{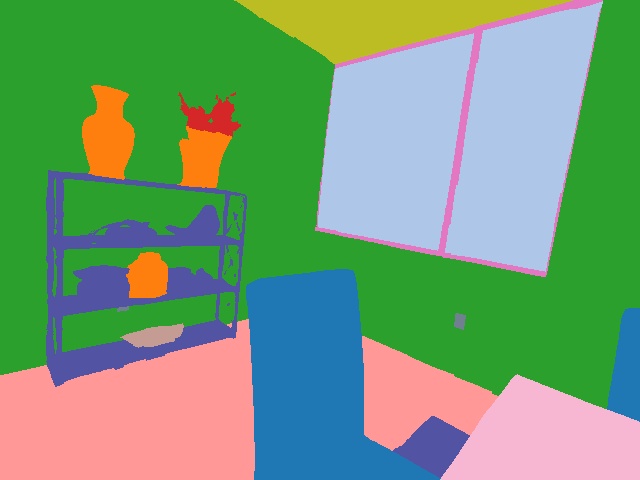} \\
\includegraphics[width=3.0cm]{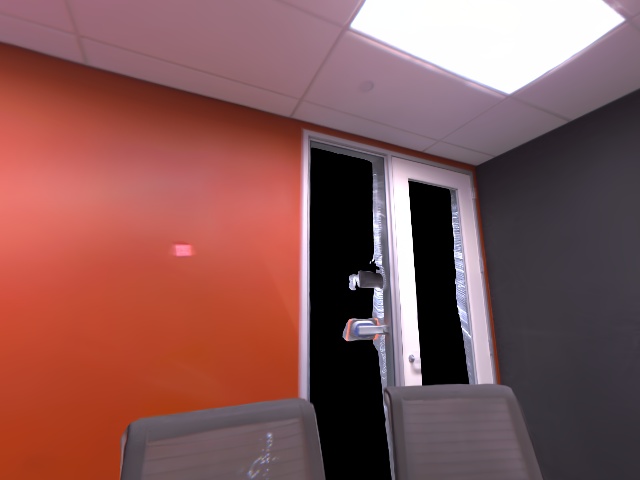} & \includegraphics[width=3.0cm]{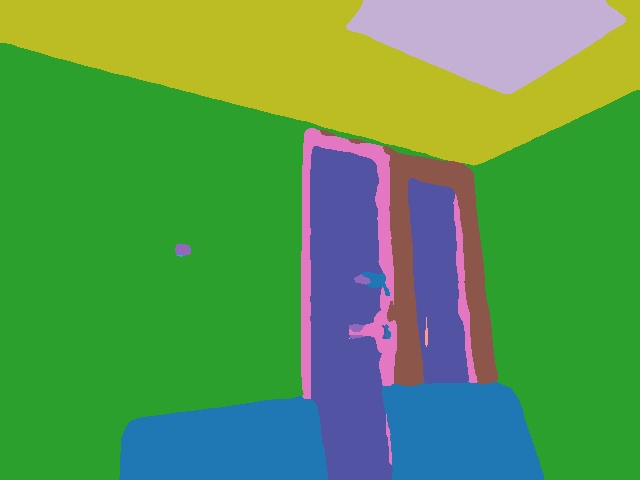} & \includegraphics[width=3.0cm]{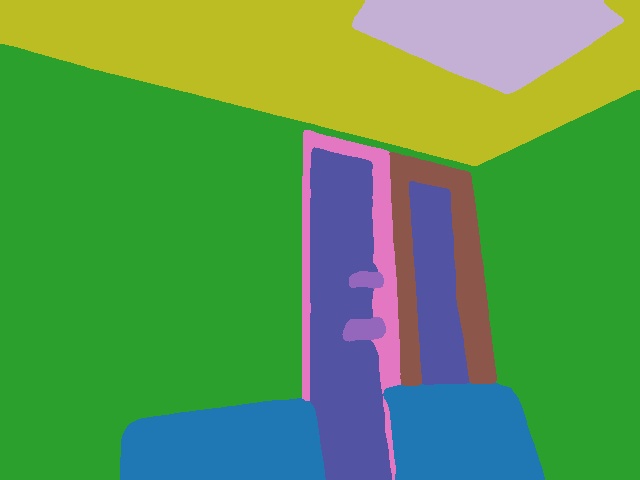} & \includegraphics[width=3.0cm]{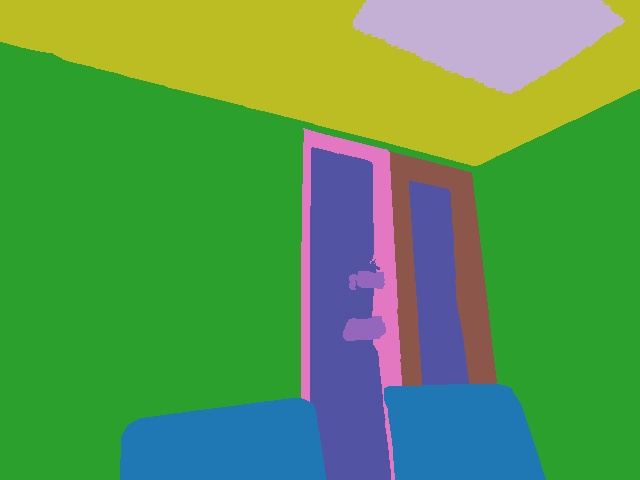} \\
\includegraphics[width=3.0cm]{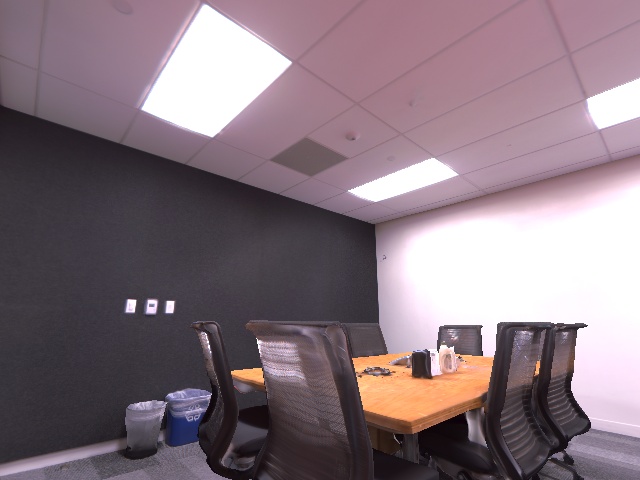} & \includegraphics[width=3.0cm]{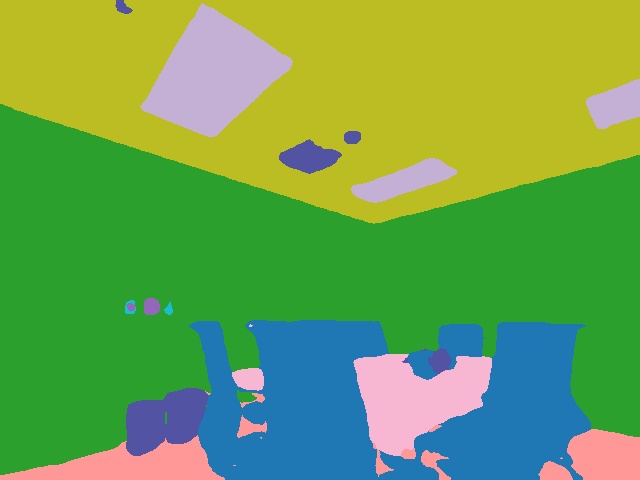} & \includegraphics[width=3.0cm]{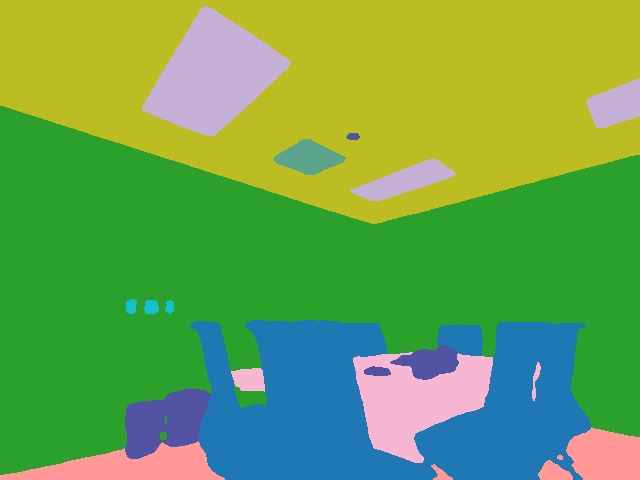} & \includegraphics[width=3.0cm]{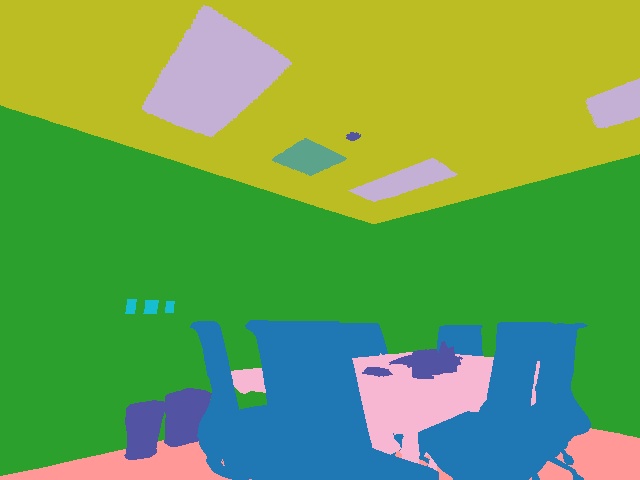} \\
\includegraphics[width=3.0cm]{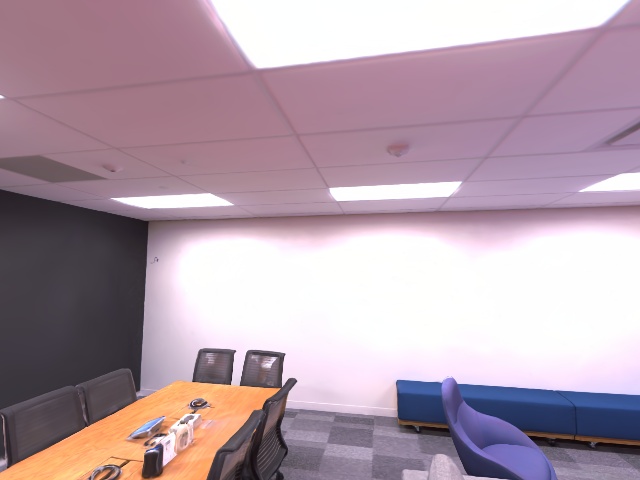} & \includegraphics[width=3.0cm]{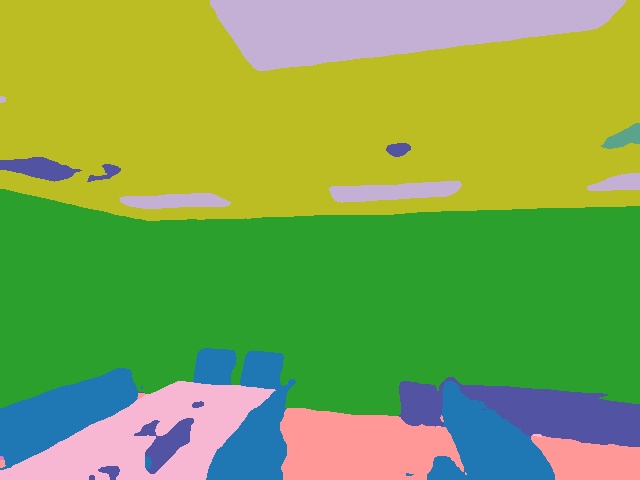} & \includegraphics[width=3.0cm]{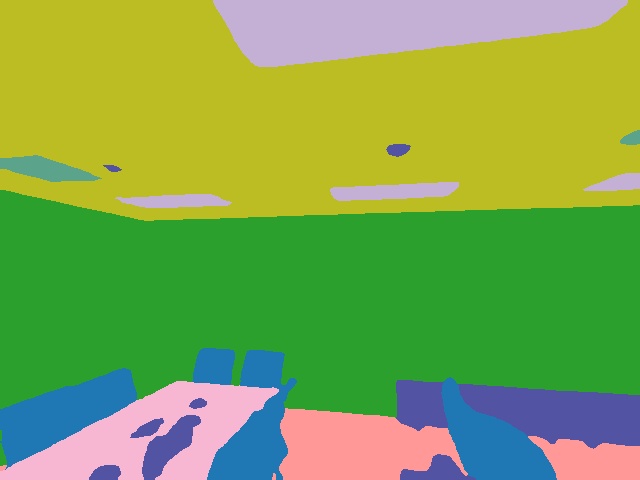} & \includegraphics[width=3.0cm]{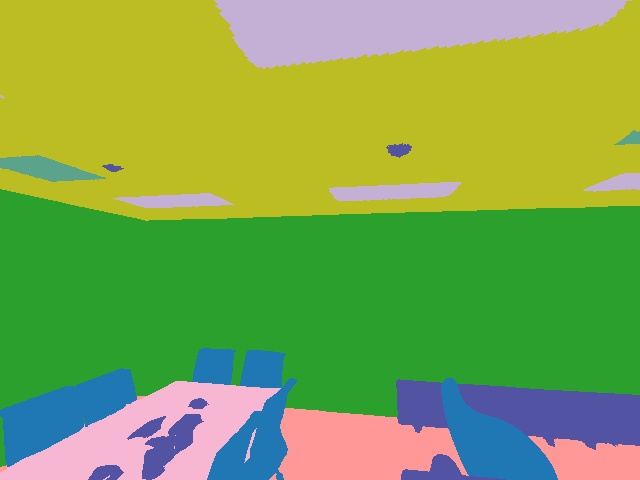} \\
Rendering & Sem. Prediction & Finetuned & GT \\
\end{tabular}
\caption{Additional qualitative results of RT-GS2 on the Replica~\cite{straub2019replica} dataset.}
\label{tab:qual_replica}
\end{figure}

\begin{figure}[H]
\centering
\begin{tabular}{>{\centering\arraybackslash}m{2.7cm}>{\centering\arraybackslash}m{2.7cm}>{\centering\arraybackslash}m{2.7cm}>{\centering\arraybackslash}m{2.7cm}}
\includegraphics[width=3.0cm]{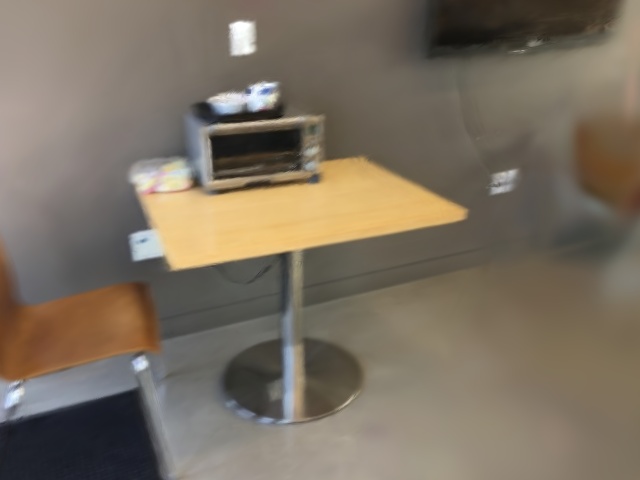} & \includegraphics[width=3.0cm]{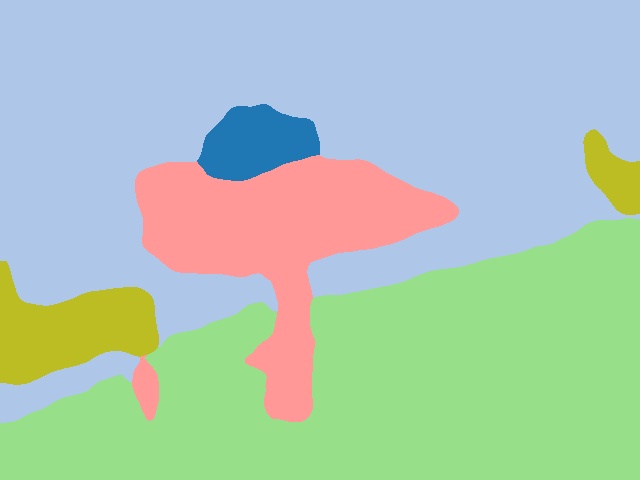} & \includegraphics[width=3.0cm]{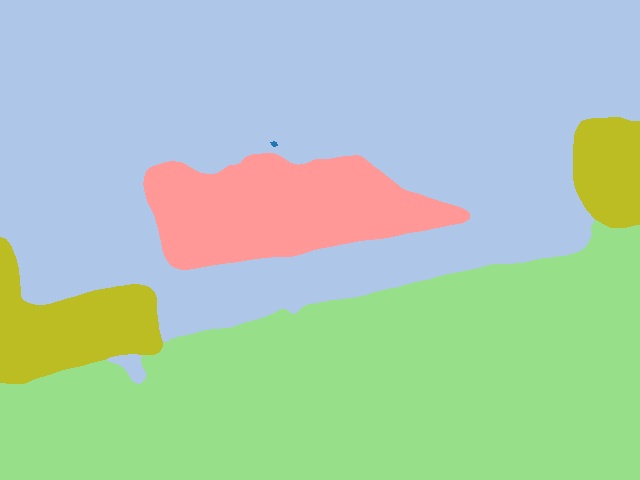} & \includegraphics[width=3.0cm]{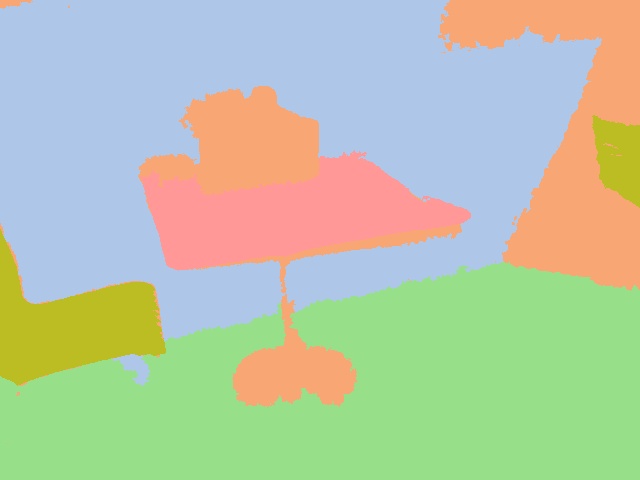}\\
\includegraphics[width=3.0cm]{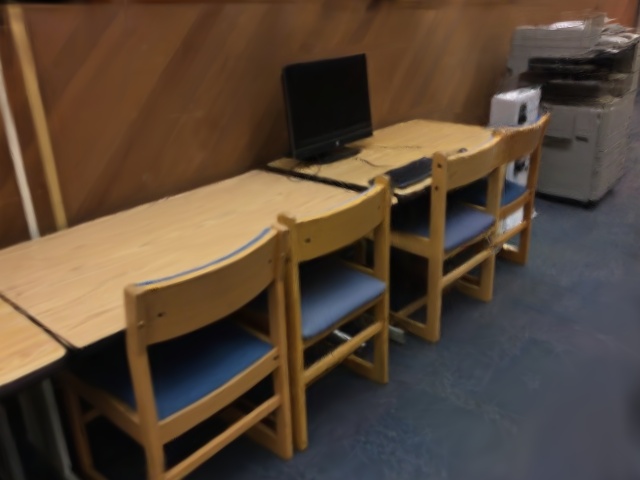} & \includegraphics[width=3.0cm]{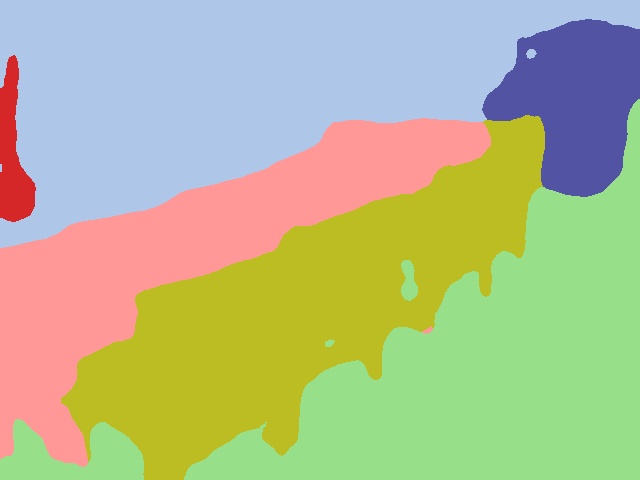} & \includegraphics[width=3.0cm]{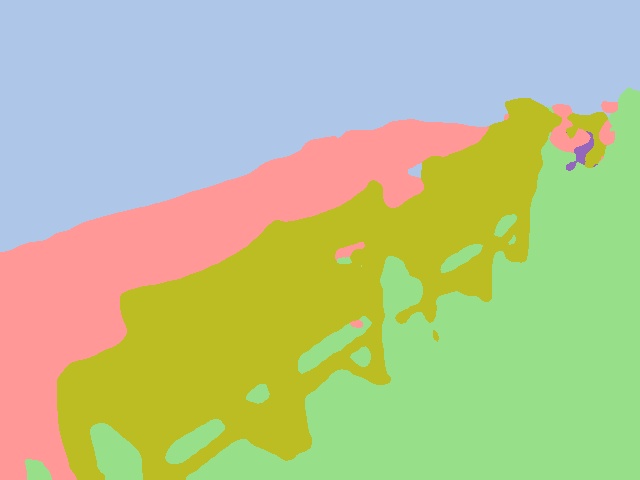} & \includegraphics[width=3.0cm]{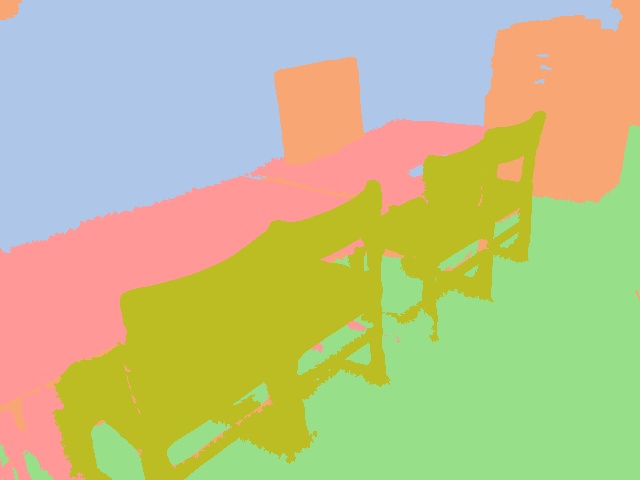} \\
\includegraphics[width=3.0cm]{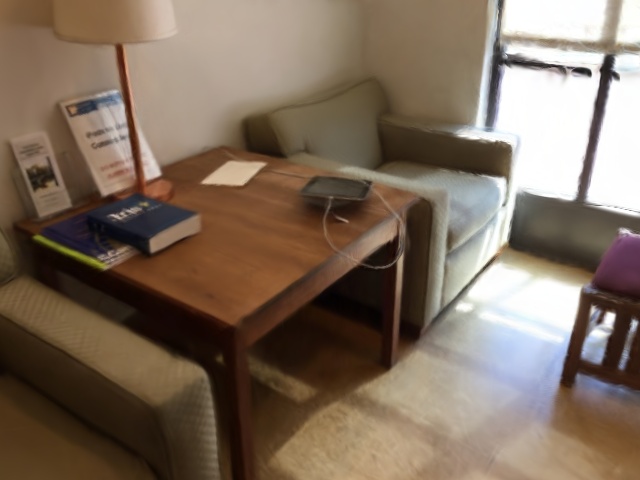} & \includegraphics[width=3.0cm]{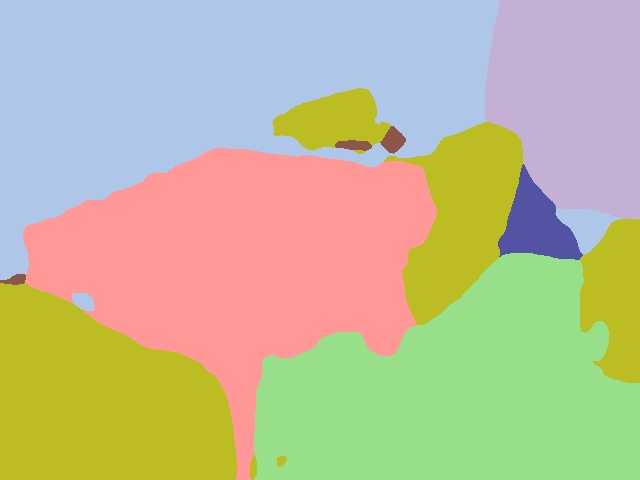} & \includegraphics[width=3.0cm]{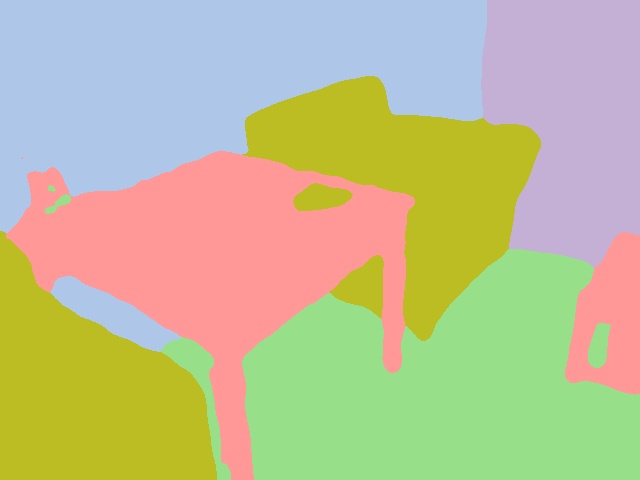} & \includegraphics[width=3.0cm]{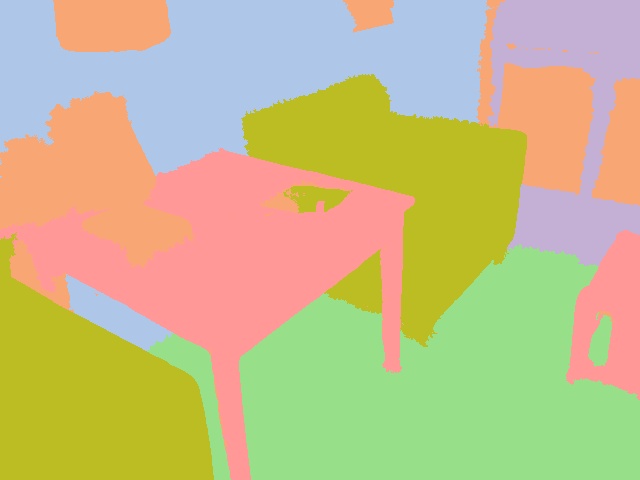} \\
\includegraphics[width=3.0cm]{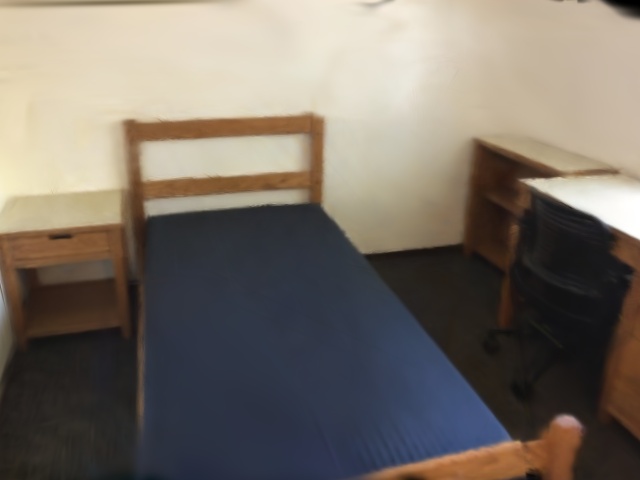} & \includegraphics[width=3.0cm]{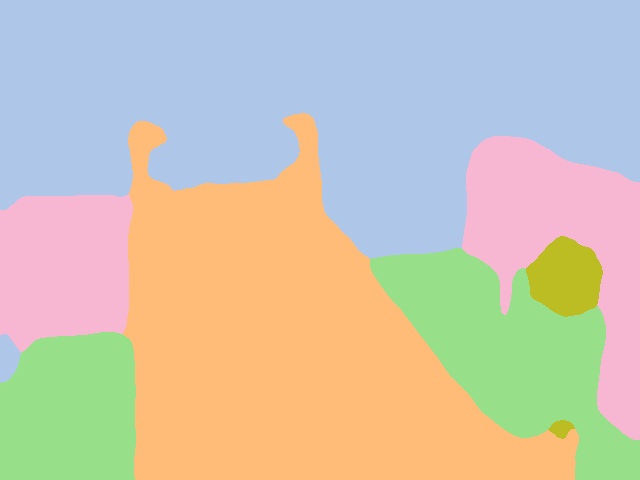} & \includegraphics[width=3.0cm]{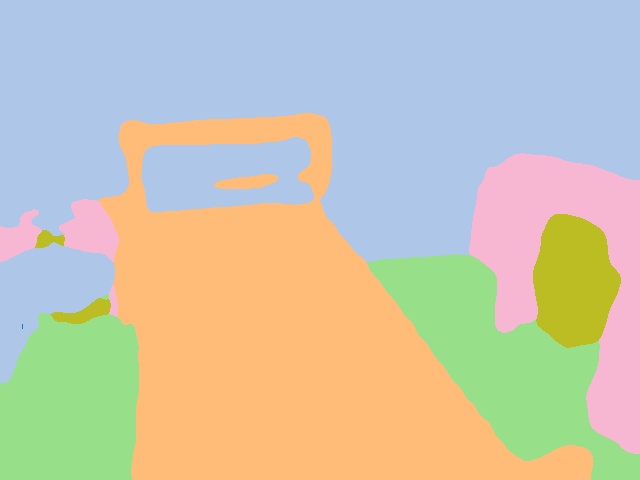} & \includegraphics[width=3.0cm]{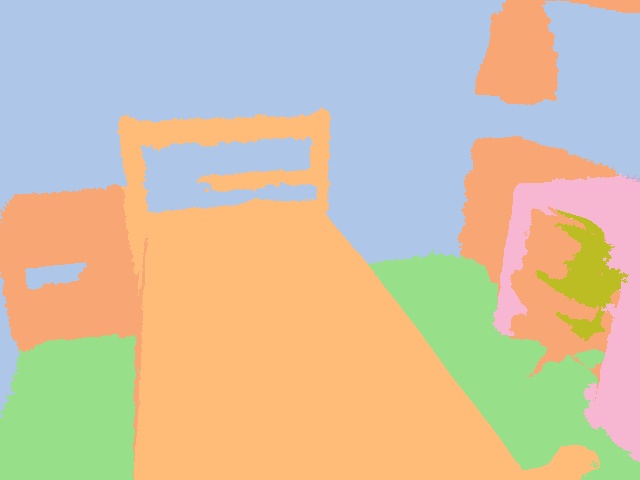} \\
\includegraphics[width=3.0cm]{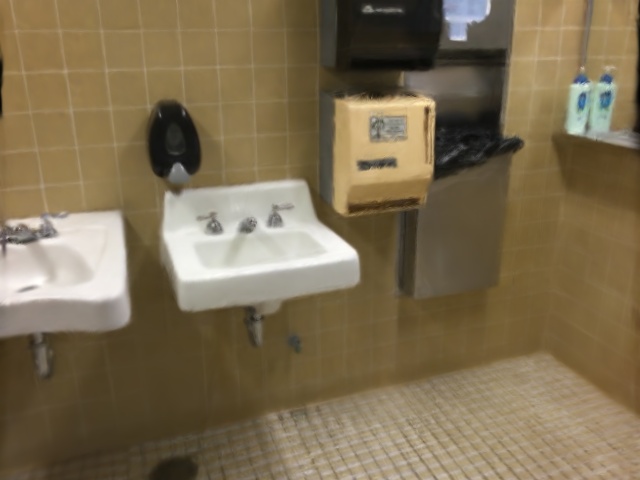} & \includegraphics[width=3.0cm]{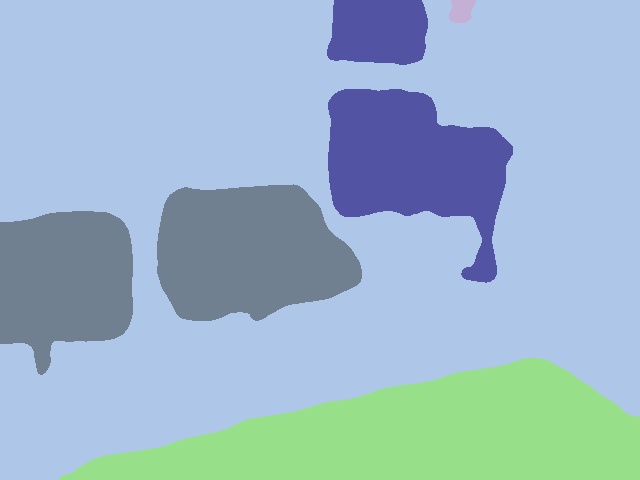} & \includegraphics[width=3.0cm]{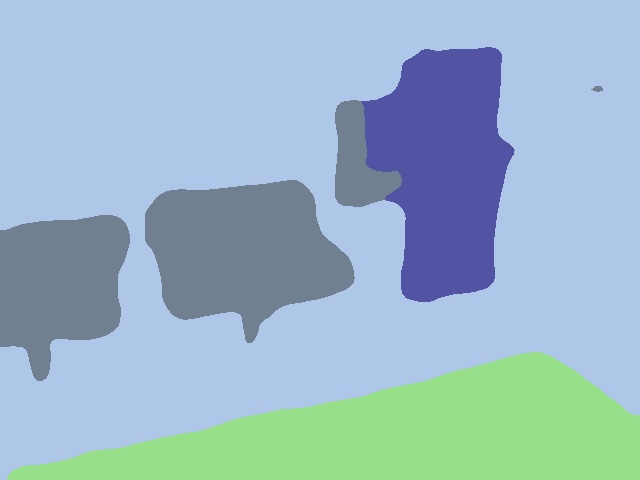} & \includegraphics[width=3.0cm]{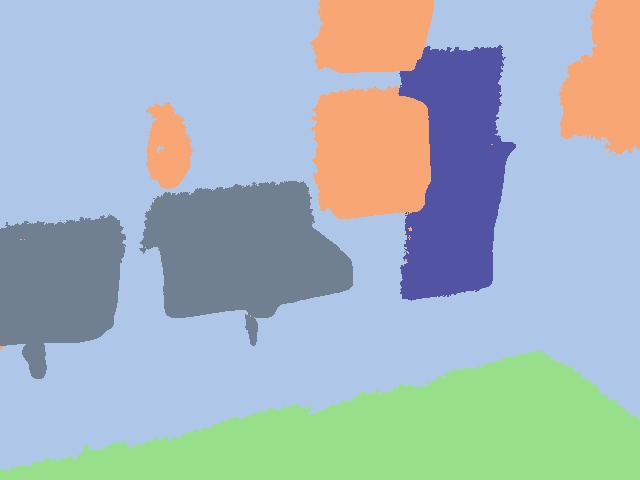} \\
Rendering & Sem. Prediction & Finetuned & GT \\
\end{tabular}
\caption{Additional qualitative results of RT-GS2 on the ScanNet~\cite{dai2017scannet} dataset. We point out that the peach orange color is the unannotated class, which is frequently present in the ScanNet dataset.}
\label{tab:qual_scannet}
\end{figure}

\begin{figure}[H]
\centering
\begin{tabular}{>{\centering\arraybackslash}m{2.7cm}>{\centering\arraybackslash}m{2.7cm}>{\centering\arraybackslash}m{2.7cm}>{\centering\arraybackslash}m{2.7cm}}
\includegraphics[width=3.0cm]{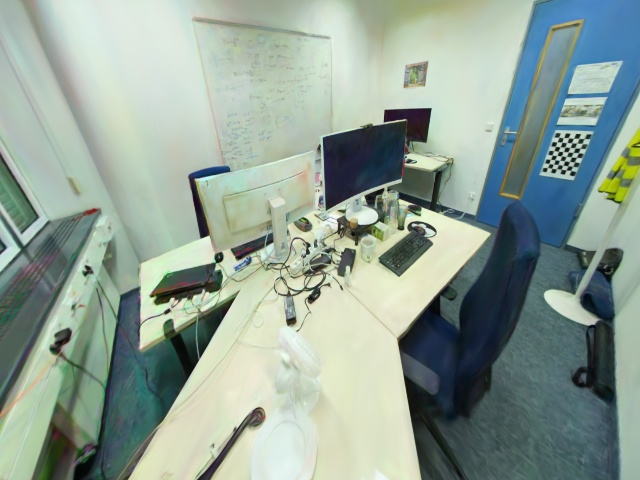} & \includegraphics[width=3.0cm]{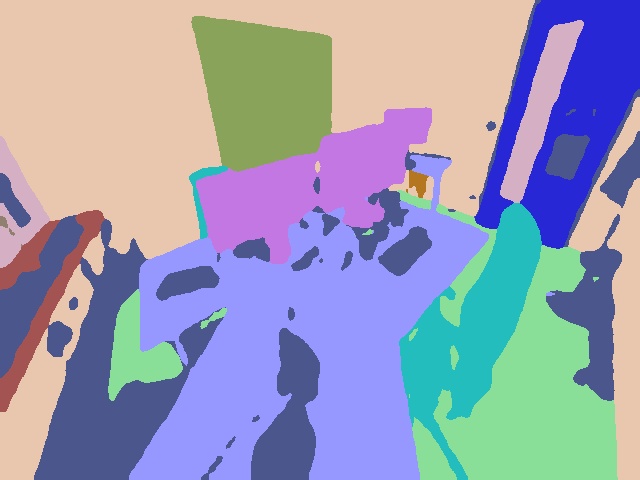} & \includegraphics[width=3.0cm]{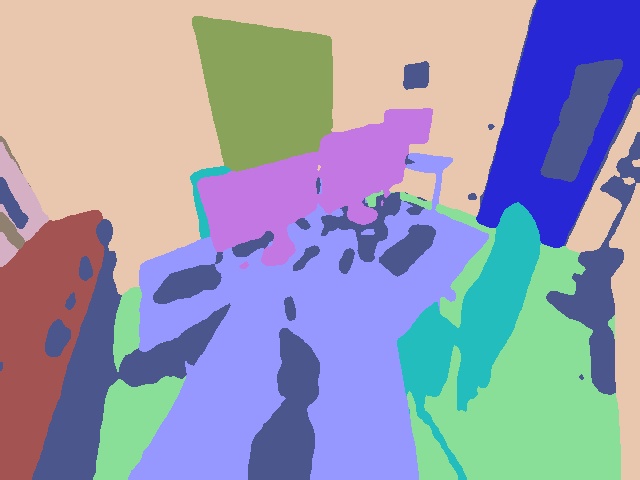} & \includegraphics[width=3.0cm]{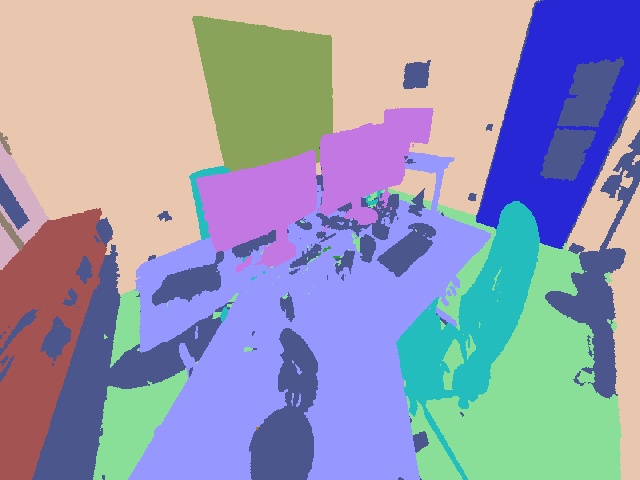} \\
\vspace{-0.3em}
\includegraphics[width=3.0cm]{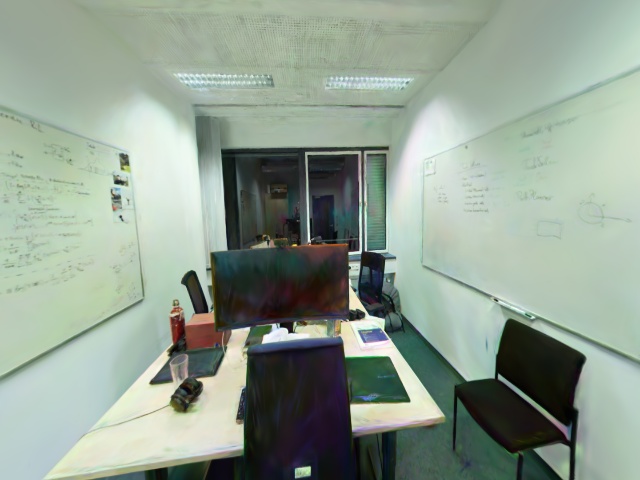} & \includegraphics[width=3.0cm]{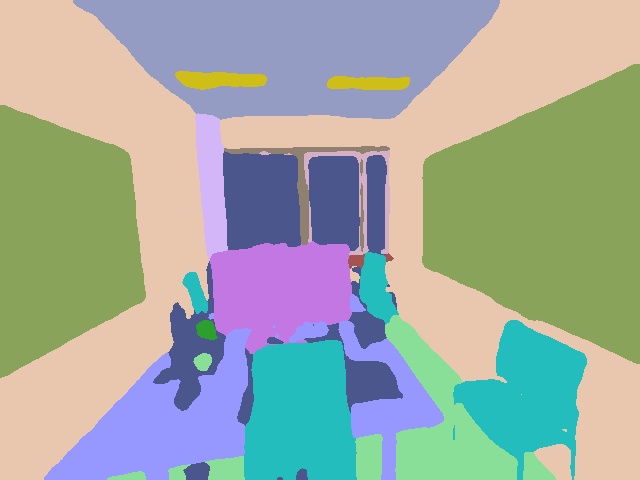} & \includegraphics[width=3.0cm]{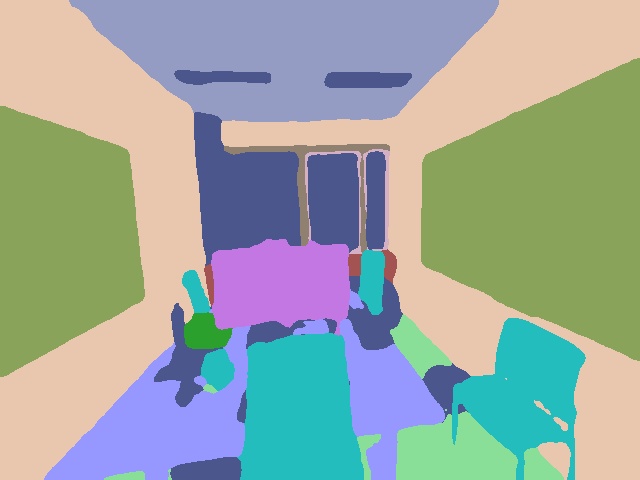} & \includegraphics[width=3.0cm]{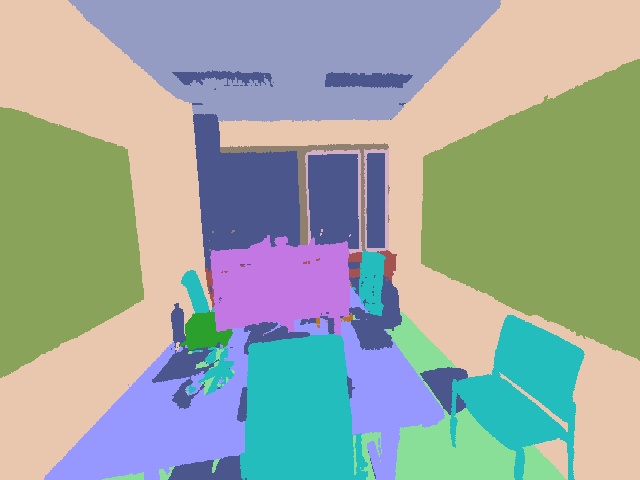} \\
\vspace{-0.3em}
\includegraphics[width=3.0cm]{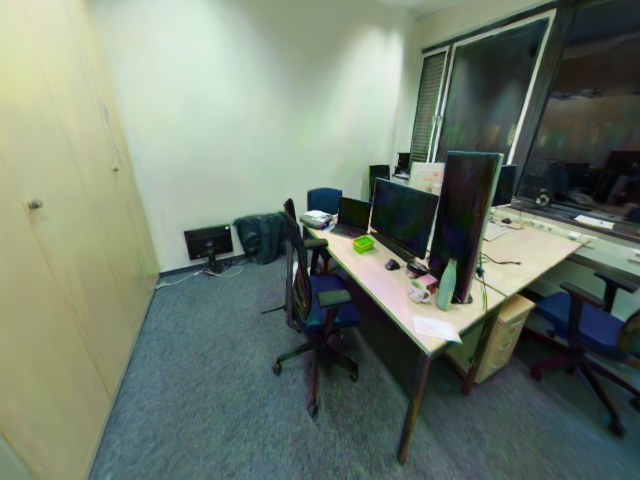} & \includegraphics[width=3.0cm]{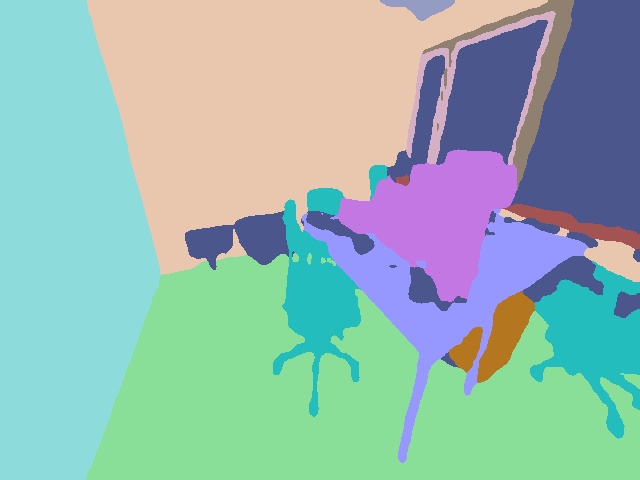} & \includegraphics[width=3.0cm]{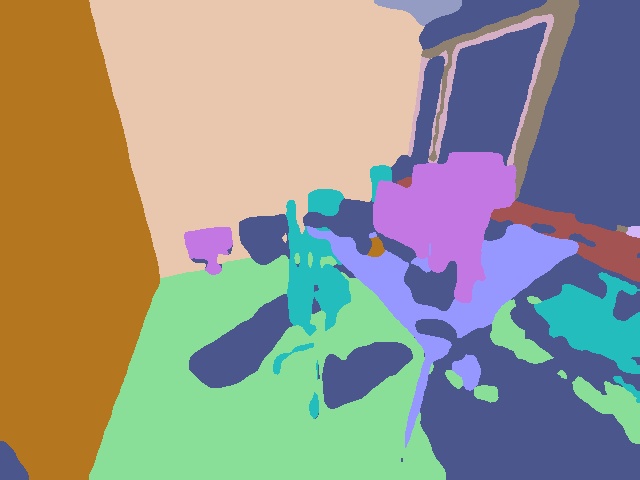} & \includegraphics[width=3.0cm]{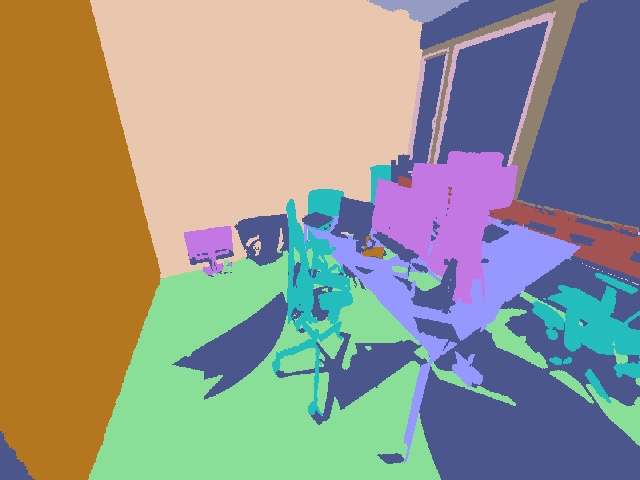} \\
\vspace{-0.3em}
\includegraphics[width=3.0cm]{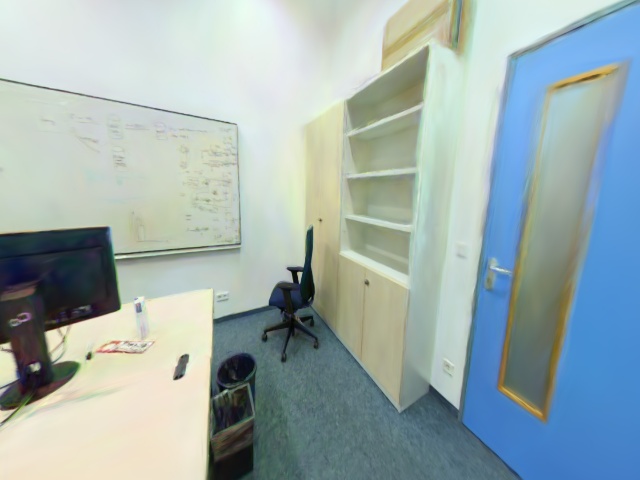} & \includegraphics[width=3.0cm]{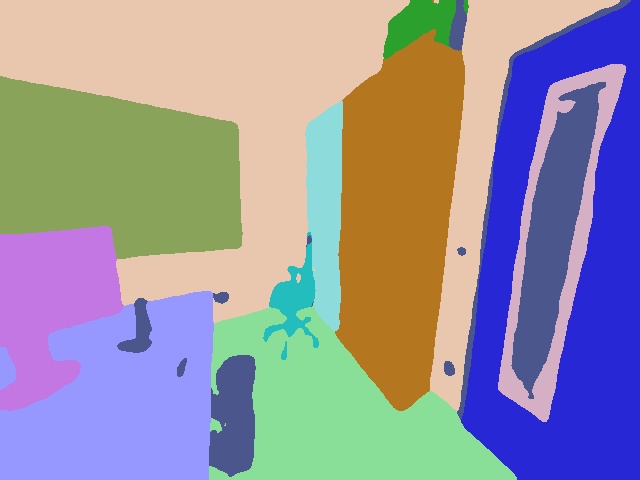} & \includegraphics[width=3.0cm]{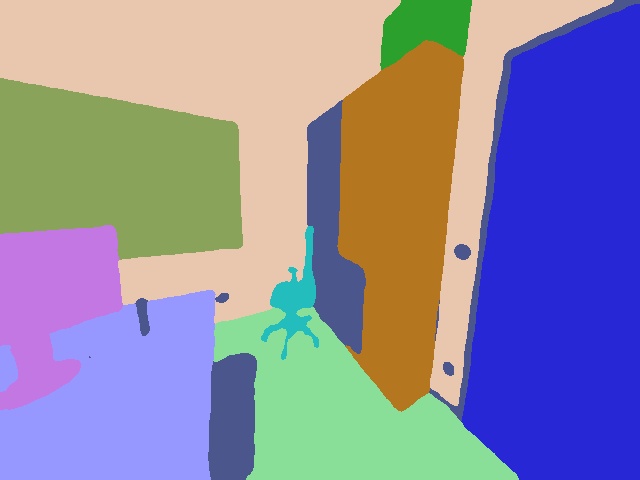} & \includegraphics[width=3.0cm]{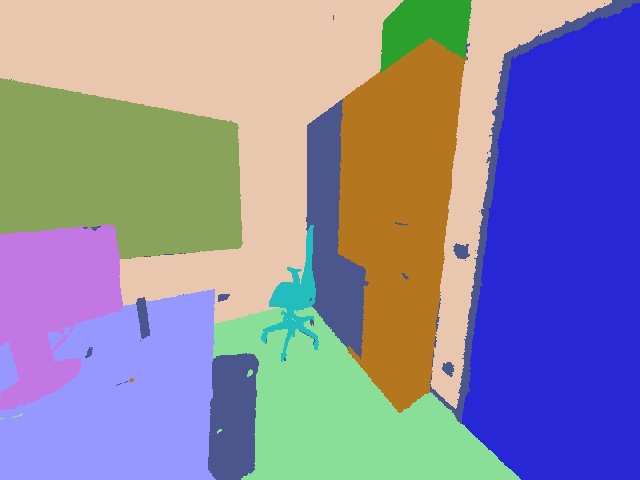} \\
\vspace{-0.3em}
\includegraphics[width=3.0cm]{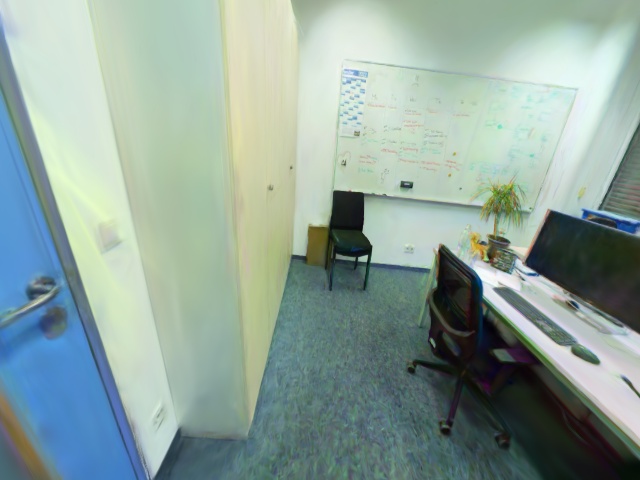} & \includegraphics[width=3.0cm]{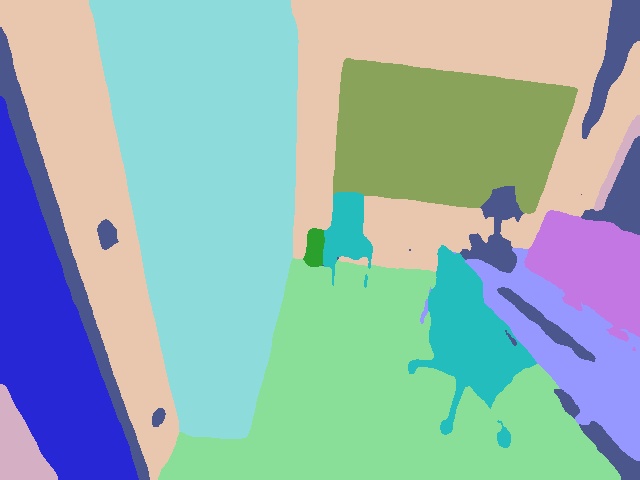} & \includegraphics[width=3.0cm]{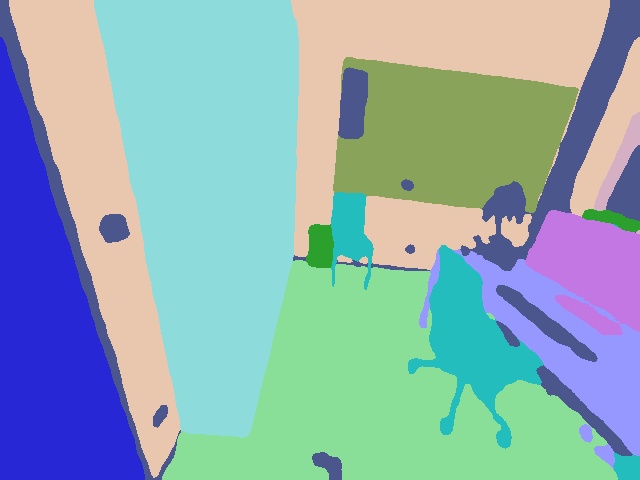} & \includegraphics[width=3.0cm]{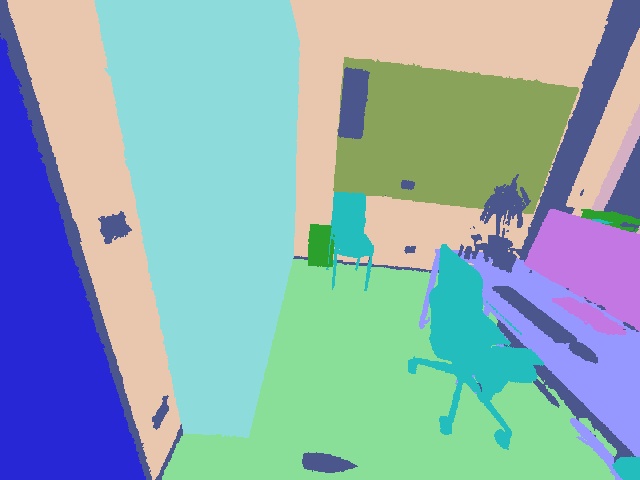} \\
Rendering & Prediction & Finetuned & GT \\
\end{tabular}
\caption{Qualitative results on ScanNet++~\cite{yeshwanth2023scannet++} on the first five test scenes. We point out that dark purple color represents the unannotated and other classes in the ScanNet++ dataset.}
\label{tab:scanetpp_viz1}
\end{figure}

\begin{figure}[H]
\centering
\begin{tabular}{>{\centering\arraybackslash}m{2.7cm}>{\centering\arraybackslash}m{2.7cm}>{\centering\arraybackslash}m{2.7cm}>{\centering\arraybackslash}m{2.7cm}}
\includegraphics[width=3.0cm]{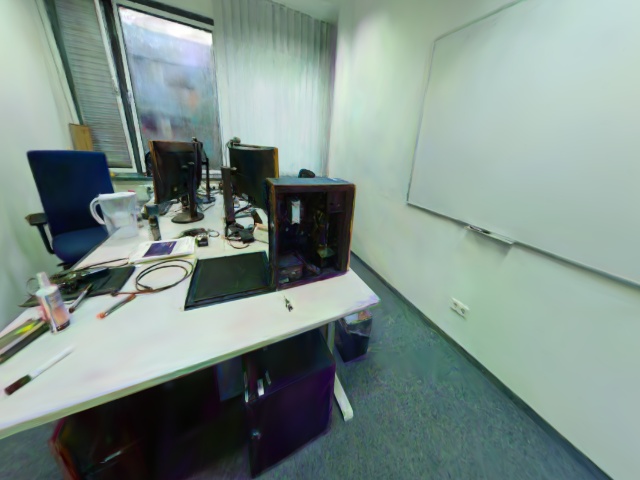} & \includegraphics[width=3.0cm]{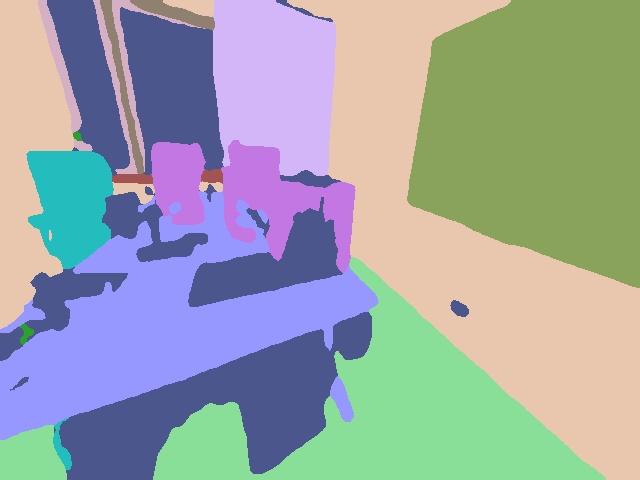} & \includegraphics[width=3.0cm]{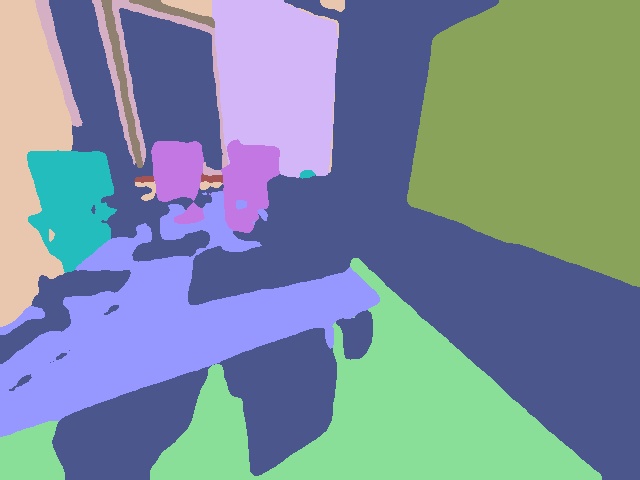} & \includegraphics[width=3.0cm]{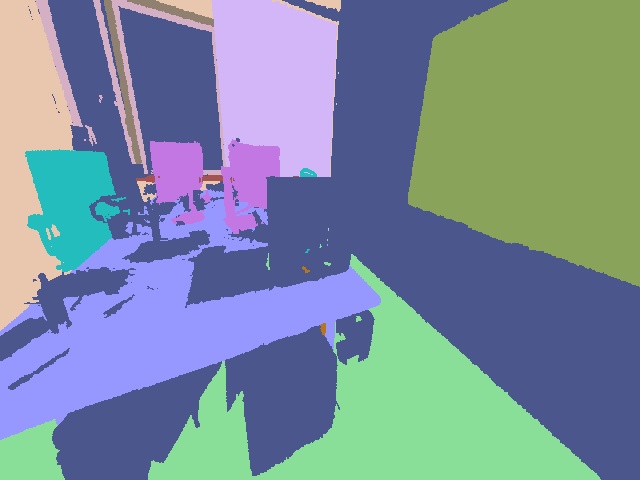} \\
\vspace{-0.3em}
\includegraphics[width=3.0cm]{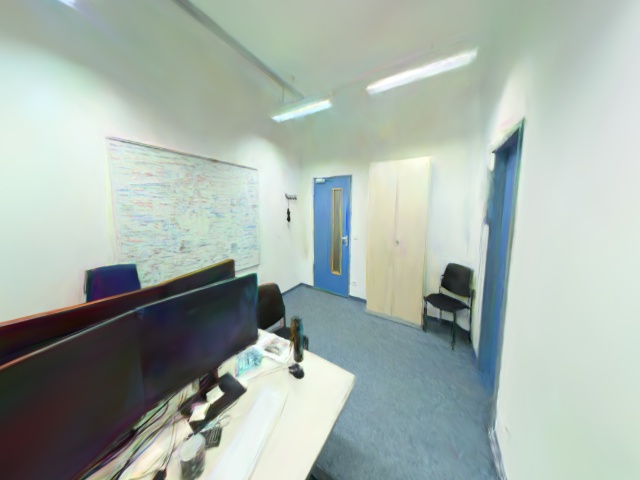} & \includegraphics[width=3.0cm]{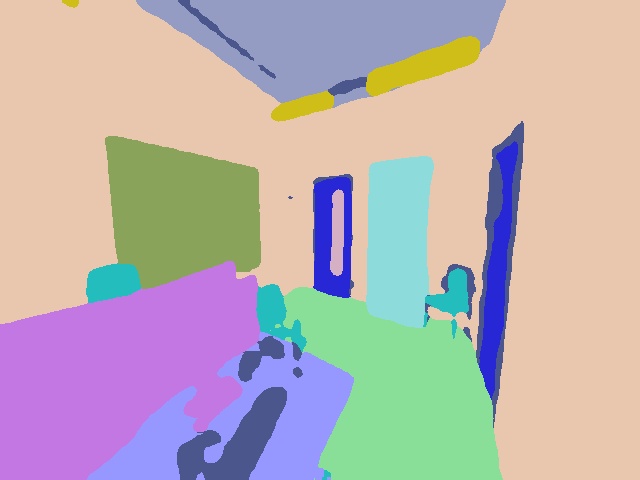} & \includegraphics[width=3.0cm]{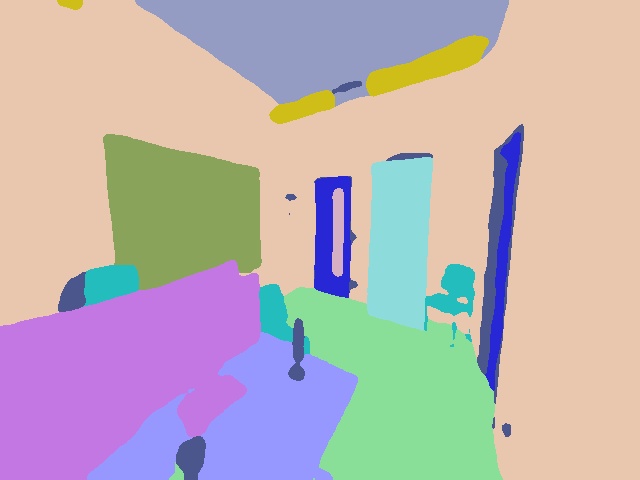} & \includegraphics[width=3.0cm]{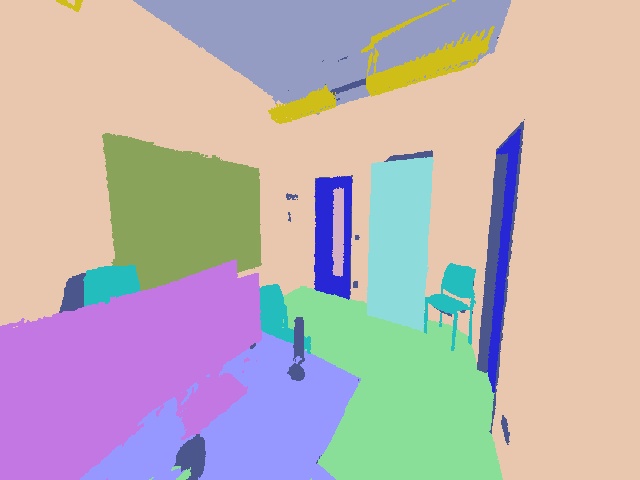} \\
\vspace{-0.3em}
\includegraphics[width=3.0cm]{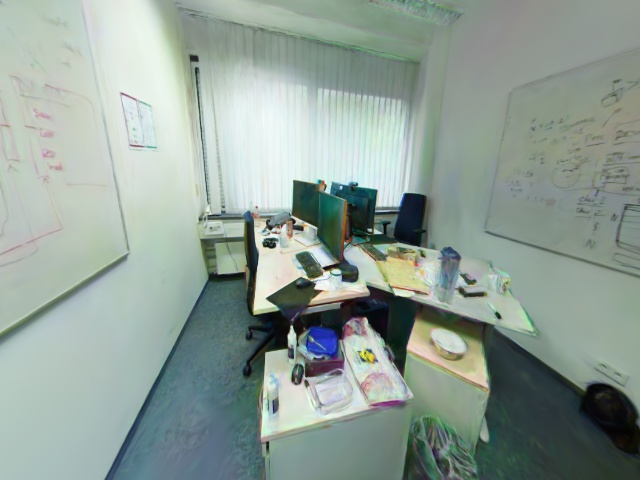} & \includegraphics[width=3.0cm]{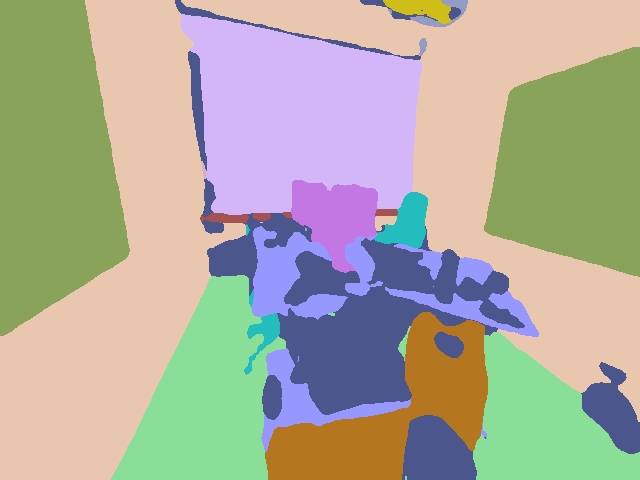} & \includegraphics[width=3.0cm]{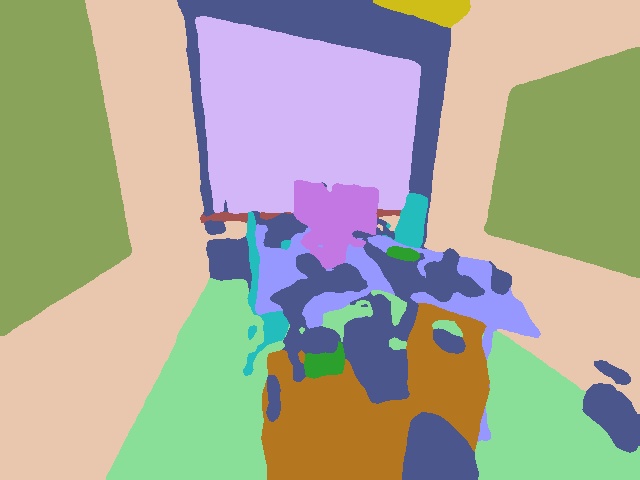} & \includegraphics[width=3.0cm]{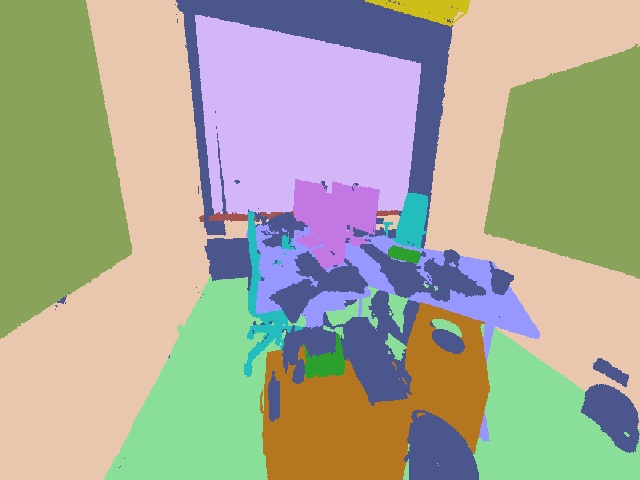} \\
\vspace{-0.3em}
\includegraphics[width=3.0cm]{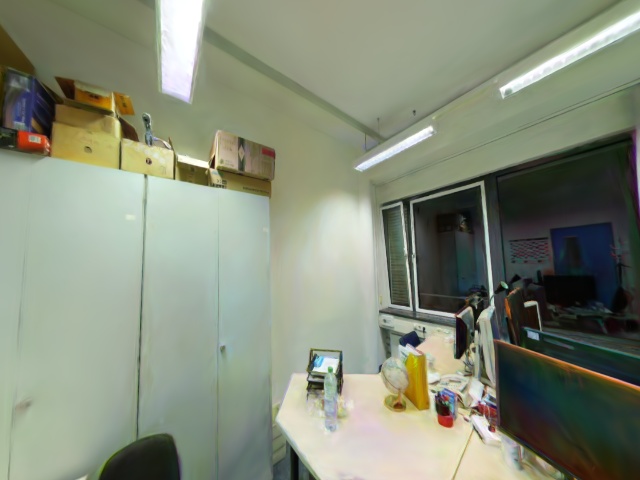} & \includegraphics[width=3.0cm]{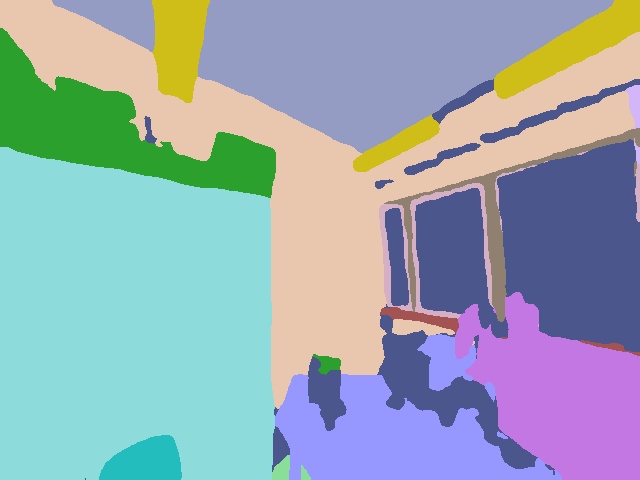} & \includegraphics[width=3.0cm]{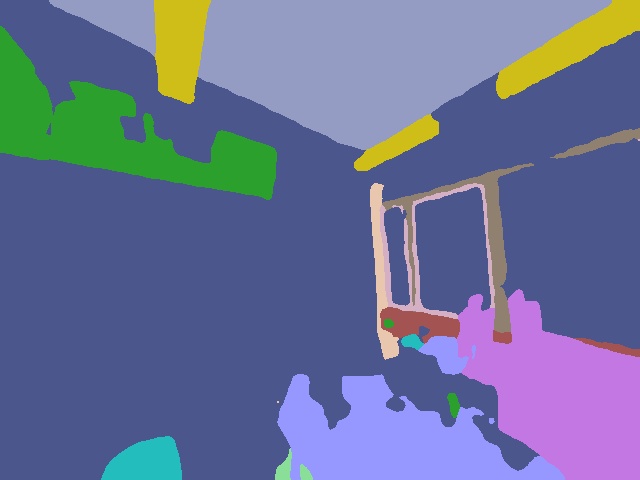} & \includegraphics[width=3.0cm]{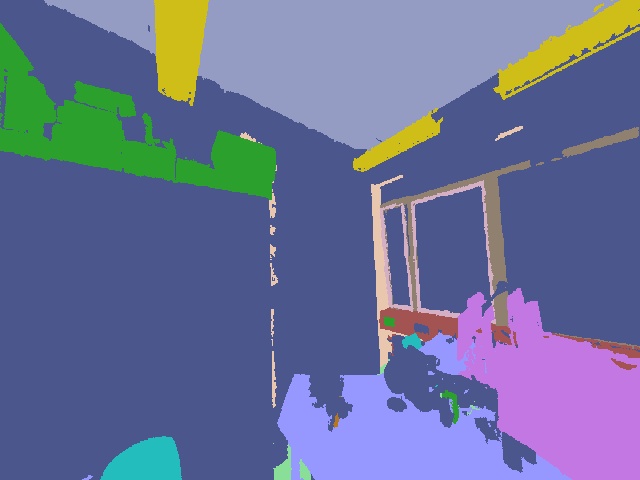} \\
\vspace{-0.3em}
\includegraphics[width=3.0cm]{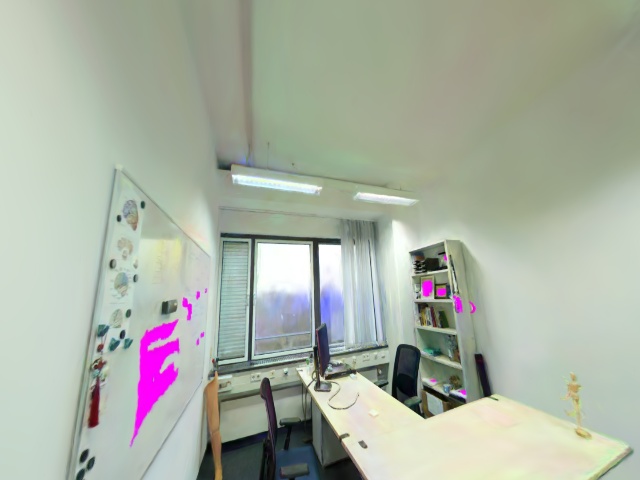} & \includegraphics[width=3.0cm]{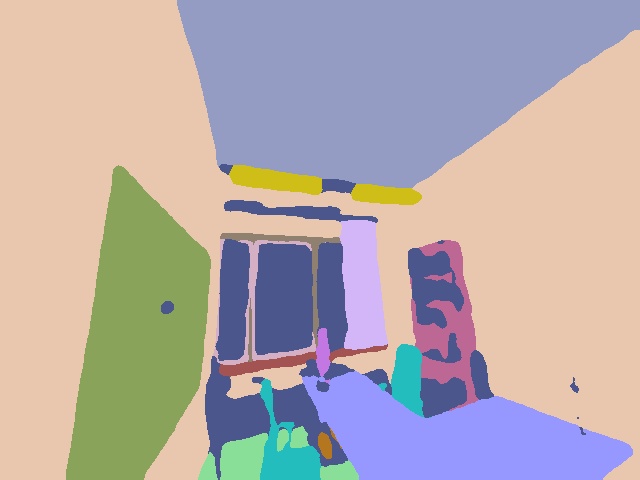} & \includegraphics[width=3.0cm]{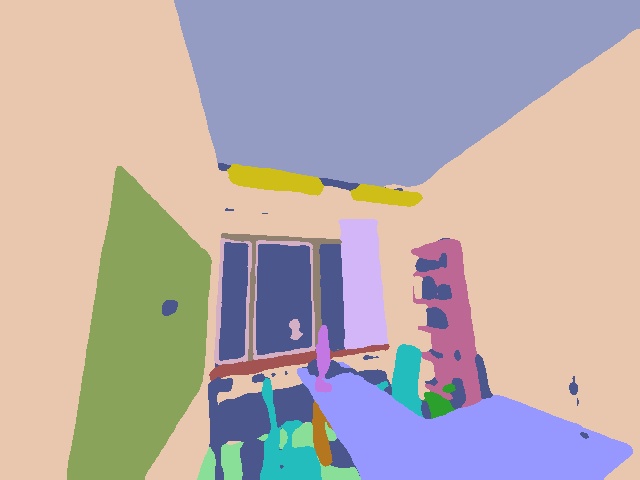} & \includegraphics[width=3.0cm]{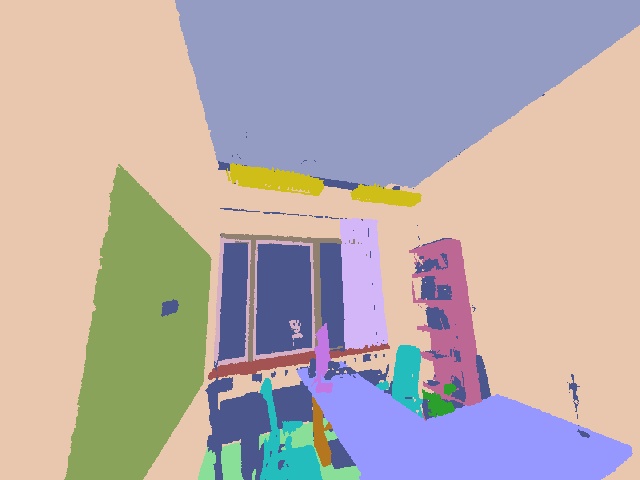} \\
Rendering & Prediction & Finetuned & GT \\
\end{tabular}
\caption{Qualitative results on ScanNet++~\cite{yeshwanth2023scannet++} on the last five test scenes. We point out that dark purple color represents the unannotated and other classes in the ScanNet++ dataset.}
\label{tab:scanetpp_viz2}
\end{figure}

In this section, we show additional qualitative visualizations on all three datasets. Figure \ref{tab:qual_replica} and Figure \ref{tab:qual_scannet} contain additional visualizations from Replica~\cite{straub2019replica} and ScanNet~\cite{dai2017scannet}, respectively. Figure \ref{tab:scanetpp_viz1} and Figure \ref{tab:scanetpp_viz2} present extensive visualizations on ScanNet++~\cite{yeshwanth2023scannet++} for all 10 test scenes on which experiments were conducted.

\newpage

\end{document}